\documentclass[letterpaper,10pt,conference]{ieeeconf}

\IEEEoverridecommandlockouts
\usepackage[T1]{fontenc}
\usepackage[utf8]{inputenc}

\usepackage{amsmath,amssymb,amsfonts}
\usepackage{bm}
\usepackage{mathtools}

\usepackage{graphicx}
\usepackage{subcaption}

\usepackage{booktabs}
\usepackage{multirow}
\usepackage{array}
\usepackage{tabularx}
\usepackage{threeparttable}
\usepackage{makecell}
\usepackage[table]{xcolor}
\usepackage{siunitx}

\usepackage{cite}
\usepackage{url}

\usepackage{algorithm}
\usepackage{algpseudocode}

\usepackage{microtype}
\usepackage{comment}

\usepackage{hyperref}
\hypersetup{
    hidelinks,
    colorlinks=false,
    pdfborder={0 0 0}
}

\newcolumntype{Y}{>{\centering\arraybackslash}X}

\newcolumntype{L}{>{\raggedright\arraybackslash}X}

\definecolor{headergray}{gray}{0.88}
\definecolor{sectiongray}{gray}{0.93}
\definecolor{rowgray}{gray}{0.97}

\title{\LARGE \bf
Hierarchical Topology-Aware Planning and Control of
Underwater Vehicle--Manipulator Systems in Confined Environments
}

\author{Mohamed Abdelwahab$^{1,2}$, Ruggero Carli$^{1}$,  Damiano Varagnolo$^{3}$ and Alberto Dalla Libera$^{1}$%
\thanks{This work was funded by the European Union--NextGenerationEU under the National Recovery and Resilience Plan (NRRP), CUP D93C23000450005.}%
\thanks{$^{1}$Mohamed Abdelwahab, Ruggero Carli, and Alberto Dalla Libera are with the Department of Information Engineering, University of Padova, Via Gradenigo 6/B, 35131 Padova, Italy.}%
\thanks{$^{2}$Mohamed Abdelwahab is also with the Department of Electrical and Information Engineering, Polytechnic University of Bari, Via Orabona 4, 70125 Bari, Italy.
        {\tt\small m.mohamed@phd.poliba.it}}%
\thanks{$^{3}$Damiano Varagnolo is with the Department of Engineering Cybernetics, Norwegian University of Science and Technology, O. S. Bragstads plass 2D, 7034 Trondheim, Norway.}%
}

\begin{document}
\maketitle
\thispagestyle{empty}
\pagestyle{empty}

\begin{abstract}
This paper addresses autonomous intervention with an underwater vehicle--manipulator system (UVMS) in confined, cluttered, and partially known environments, where poor maneuverability, narrow passages, and uncertain execution may cause the robot to enter unrecoverable regions. We propose \emph{MANTA}, a three-layer hierarchical planning-and-control framework that couples passage accessibility, manipulation feasibility, and closed-loop execution. The first layer performs global connectivity reasoning in a conservative reduced base space to extract traversable corridor candidates toward the task region. The second layer refines each candidate corridor by jointly optimizing the continuous base motion and arm trajectory, producing a collision-free base--arm trajectory. The third layer learns a reach-and-hold base policy using Gaussian-process model-based reinforcement learning (MBRL) through MC-PILCO, enabling trajectory tracking and station keeping at the planned manipulation state. During execution, the framework monitors map updates and can trigger recovery and route repair when the active passage becomes infeasible. MANTA is evaluated in confined UVMS planning and closed-loop tracking experiments. Across 120 matched planning queries, it achieves higher task success than full-state sampling-based baselines while producing larger clearance margins and lower arm motion. The learned MC-PILCO policy further reduces position and yaw tracking errors on both training and unseen tube-like references. These results show MANTA as a structured and data-efficient framework for safe autonomous underwater intervention in caves, tubes, and cluttered subsea structures.
\end{abstract}


\section{Introduction}

\label{sec:introduction}

Underwater intervention is moving beyond open-water inspection and station keeping toward operations that require physical interaction in geometrically constrained environments. Representative examples include valve turning, tool deployment, inspection behind structural members, and manipulation inside recessed or partially enclosed workspaces. In these settings, an underwater vehicle--manipulator system (UVMS) must not only reach the vicinity of the target. It must reach it through cluttered free space, preserve clearance for the vehicle and arm, and arrive at a base pose from which the manipulator can operate safely and effectively. The navigation and manipulation requirements are therefore coupled by the geometry of the environment, especially in cave-like, tube-like, and cluttered subsea structures~\cite{Sivcev2018Review,Simetti2020AutonomousIntervention,Sun2024UnderwaterRobots}.

The difficulty is threefold. First, the vehicle must identify an admissible passage through branched and narrow free space. This is a global accessibility problem, because locally attractive target neighborhoods may be unreachable through the available corridors. Second, the selected terminal base pose must support the manipulation task. Collision-free proximity to the target is insufficient if the arm reaches the target near a joint limit, with poor task-oriented dexterity, or with a configuration that is sensitive to small disturbances. Third, the usefulness of the planned operating state depends on the closed-loop execution accuracy of the vehicle. In confined underwater environments, a nominal path that is safe under ideal kinematics may become unsafe when tracking errors, hydrodynamic disturbances, and vehicle--arm coupling are taken into account~\cite{Schjolberg1994UVMS,Ding2021StationKeeping,Tijjani2022TrackingSurvey}.

\begin{figure*}[!ht]
    \centering
    \includegraphics[width=\textwidth]{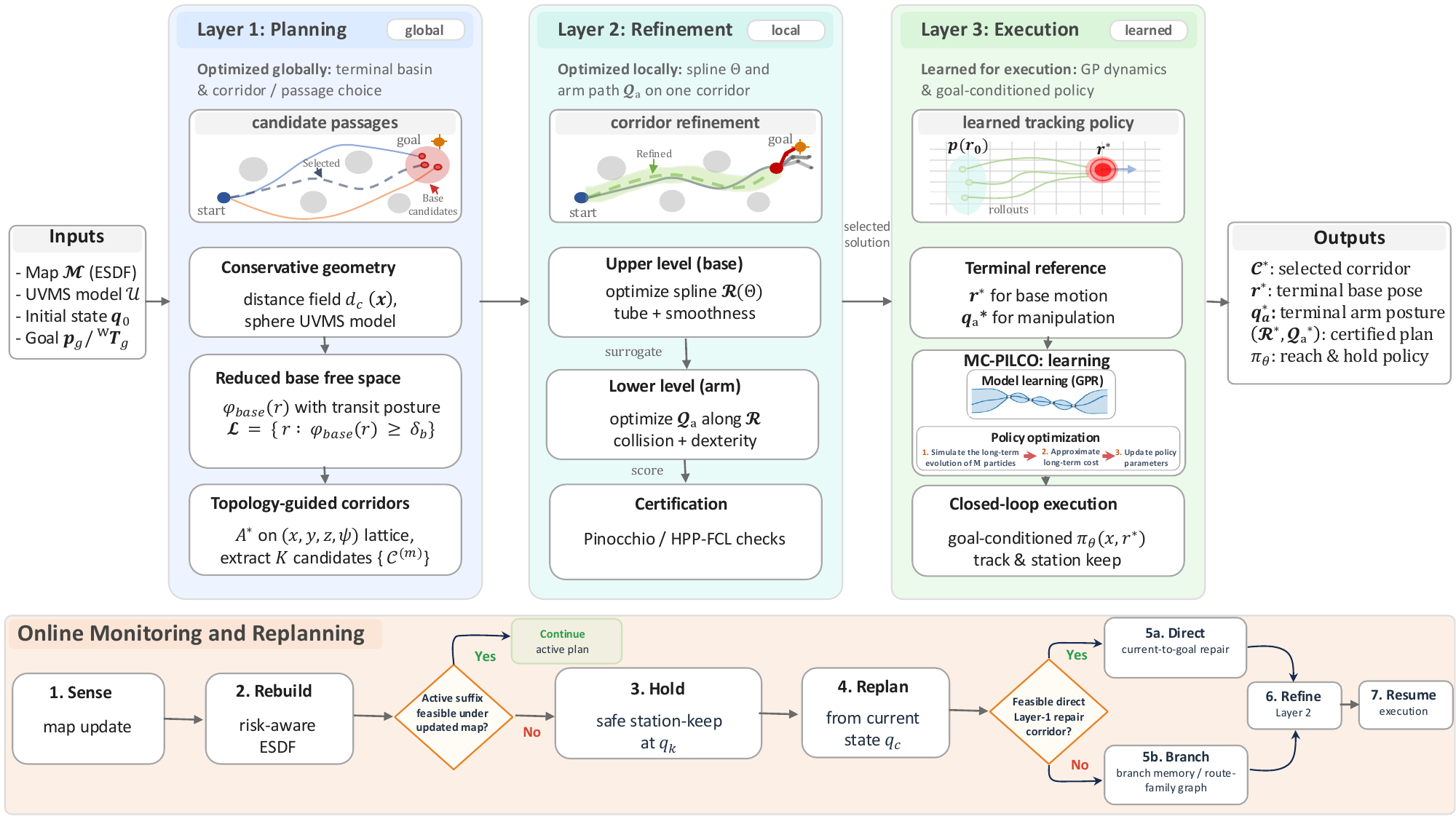}
    \caption{An overview of the proposed method.}
    \label{fig:method_overview}
\end{figure*}

Existing UVMS work has made substantial progress on local intervention control, redundancy resolution, visual servoing, obstacle avoidance, and task execution near a given worksite~\cite{Prats2014FloatingManipulation,Cieslak2015PanelOperation,SIMETTI201840,Santos2023VisualServoing,cieslak2020practical}. These methods are essential for autonomous intervention, but they typically assume that the vehicle has already reached a suitable neighborhood of the task. They therefore do not explicitly answer which confined passage should be traversed before the manipulation phase begins. A complementary literature on manipulation-aware base placement addresses where a mobile manipulator should stand for a given task, using reachability, dexterity, inverse-reachability, or comfort-region criteria~\cite{ASOKAN2005747,Sotiropoulos2011Docking,Vahrenkamp2013ReachabilityInversion,Chen2019DexterousGrasping}. These methods are also relevant, but they are often pointwise: they evaluate candidate operating poses near the target without resolving how the full vehicle--manipulator system can reach such poses through a narrow or branched environment.

Global planning provides the missing accessibility layer. Task-space constrained planning and sampling-based methods provide general tools for constrained motion
generation~\cite{Berenson2011TSR,Kuffner2000RRTConnect,Karaman2011RRTStar}. However, narrow passages remain a classical difficulty for generic sampling-based planners, since thin feasible corridors may be poorly explored~\cite{Hsu2003BridgeTest,1545606}. Topological, homotopy-aware, skeleton-guided, and multi-representation planning methods address this issue by representing route alternatives or structured connectivity before local refinement~\cite{Bhattacharya_2010,Bhattacharya2012TopologicalConstraints,Uwacu2022SkeletonGuidance,Youakim2018LooseCoupling,Youakim2020MultirepresentationMA}. Nevertheless, generic global planning over the full UVMS configuration is computationally expensive because it couples the vehicle pose and manipulator posture over long horizons. Conversely, planning only for the vehicle base is efficient, but it does not certify arm feasibility or dexterity. A practical UVMS planner for confined intervention must therefore combine reduced global passage reasoning with local whole-body refinement.

The final missing component is execution under uncertainty. Robust, adaptive, optimal, and model-predictive controllers have been widely studied for underwater vehicles and UVMS platforms~\cite{Schjolberg1994UVMS,Ding2021StationKeeping,Tijjani2022TrackingSurvey}. However, a purely classical model-based controller can be difficult to maintain in the intervention setting considered here. Added mass, hydrodynamic damping, restoring effects, actuator characteristics, and vehicle--arm coupling are only approximately known, and may change after hardware modifications such as adding sensors, relocating payloads, or changing the manipulator configuration. A controller designed around a fixed nominal model may therefore require repeated retuning, while overly conservative robust gains can improve stability margins at the cost of tracking accuracy and responsiveness.

We address this execution problem with a data-efficient MBRL layer that learns a tracking policy from closed-loop interaction data. Gaussian-process MBRL provides a principled way to learn uncertain nonlinear dynamics and optimize feedback policies in the low-data regime~\cite{Deisenroth2011PILCO,Deisenroth2015GPControl}. In particular, we use MC-PILCO, which combines probabilistic dynamics learning with Monte Carlo policy optimization and is therefore well suited to nonlinear robotic systems where collecting large amounts of real interaction data is impractical~\cite{amadio2022mcpilco}. Nevertheless, learning a controller alone does not resolve the global geometric question of which passage should be selected, nor does it determine whether the final operating state is manipulation ready.

This paper addresses these limitations through \emph{MANTA} (\emph{Manipulator-Aware Navigation with Topology-guided Adaptation}), a hierarchical planning-and-control framework for confined UVMS intervention. The central premise is that the task should not be treated as a single monolithic planning problem. Instead, MANTA decomposes it into three coupled layers: global passage selection in a conservative reduced base space, corridor-conditioned whole-body refinement for manipulation-ready terminal-state generation, and learning-based reach-and-hold execution under uncertain underwater dynamics. This decomposition reflects the structure of the intervention problem itself: the system must access the workspace, prepare a feasible manipulation state, and execute the planned motion accurately enough to preserve the geometric margins assumed by the planner.

The main contributions are as follows:
\begin{itemize}

    \item A \emph{topology-guided manipulation-aware planning method} that
    converts high-clearance reduced corridors into certified base--arm
    trajectories with terminal task accuracy, self-collision avoidance,
    joint-limit margin, cross-section feasibility, and task-oriented dexterity.

    \item A \emph{map-versioned execution interface} that monitors the remaining
    executable suffix of the active plan under map updates, retains the plan when
    it remains valid, and invokes hold-and-repair replanning only when the
    updated map invalidates the suffix.

    \item An \emph{integration of Gaussian-process MBRL} through MC-PILCO, used to learn a goal-conditioned base policy for
    data-efficient tracking and station keeping at the planned manipulation-ready
    operating state.
\end{itemize}

The remainder of the paper is organized as follows. Section~\ref{sec:problem_formulation} formulates the access--placement--execution problem and states the modeling assumptions. Section~\ref{sec:method} presents the MANTA methodology, including global corridor extraction, corridor-conditioned refinement, online monitoring and repair, and MC-PILCO execution. Sections~\ref{sec:experimental_setup} and~\ref{sec:results} present the experimental setup and results. Section~\ref{sec:conclusion} concludes the paper.

\section{Problem Formulation}
\label{sec:problem_formulation}

\subsection{Setup}
\label{subsec:problem_setup}

We consider an UVMS composed of a floating vehicle base and an $n_a$-DOF manipulator. The full configuration is
\begin{equation*}
\mathbf q_{f}=(\mathbf q_b,\mathbf q_a),
\qquad
\mathbf q_b\in SE(3),\quad
\mathbf q_a\in\mathbb R^{n_a} .
\end{equation*}
For long-range passage reasoning, we use the reduced base state $r=[x,y,z,\psi]^\top,$ where $(x,y,z)$ is the vehicle position and $\psi$ is the yaw angle. Roll and pitch are assumed to remain close to their nominal values during passage-level search, thanks to opportune low-level controllers that are typically embedded in working-class types of remotely operated vehicles (ROVs), for example. 

Let ${}^{w}\!T_b(\mathbf r)$ denote the vehicle pose induced by the reduced state, and define the configuration induced by $(\mathbf r,\mathbf q_a)$ as
$\mathbf q = (\mathbf r,\mathbf q_a)$. The UVMS forward kinematics is $f_{\mathrm{fk}} = {}^{w}\!T_e(\mathbf q),$ where ${}^{w}\!T_e$ is the end-effector pose. The intervention task is specified by a target frame ${}^{w}\!T_g\in SE(3)$, representing either a desired tool pose or a task frame attached to the object of interest. We denote by $\mathbf p_g$ the task position extracted from ${}^{w}\!T_g$.

At map version $\eta$, the robot is given a map $\mathcal M^\eta$ of a bounded cartesian workspace containing static obstacles. The map may be incomplete at deployment and may be updated during execution. Hence, the problem is not only to identify a useful operating state near ${}^{w}\!T_g$, but also to determine whether that state remains accessible through a valid passage as the available map evolves.

\subsection{Manipulation-ready terminal states}
\label{subsec:terminal_set}

A terminal operating state must be collision-free, compatible with the terminal task, and sufficiently dexterous for the intended manipulation action. 

Let $\mathcal I_{\mathrm{env}}$ be the set of conservative robot body elements used for robot--environment clearance checking, and let $\mathcal P_{\mathrm{self}}$ be the set of body-element pairs monitored for self-collision. For each body element $i$ in $\mathcal I_{\mathrm{env}}$, $h_i(\mathbf q)$ denotes its environment clearance margin. For each pair $(i,j)\in\mathcal P_{\mathrm{self}}$, $h^{\mathrm{self}}_{ij}(\mathbf q)$ denotes the corresponding self-clearance margin. Nonnegative values indicate collision-free placement with respect to the corresponding check.

The terminal task error is measured by $\Phi_{\mathrm{task}}(\mathbf q;{}^{w}\!T_g)$, with admissible tolerance $\varepsilon_T\ge0$. Depending on the task, this term may represent a translational tool-position error or a weighted pose error. Let $M_{\mathrm{task}}(\mathbf q)$ denote a terminal manipulability index, either isotropic or task-directional, and let $\underline M>0$ be the minimum acceptable value. Summarizing, the set of conditions defining the condition of being \emph{manipulation-ready} are 
\begin{equation}
\label{eq:terminal_set_rewritten}
\mathcal X_{\mathrm{term}}
=
\left\{
\mathbf q
\;\middle|\;
\begin{aligned}
& h_i(\mathbf q)\ge 0,\;\forall i\in\mathcal{I}_{\mathrm{env}},\\
& h^{\mathrm{self}}_{ij}(\mathbf q)\ge 0,\;\forall (i,j)\in\mathcal{P}_{\mathrm{self}},\\
& \Phi_{\mathrm{task}}(\mathbf q;{}^{w}\!T_g)\le \varepsilon_T,\\
& M_{\mathrm{task}}(\mathbf q)\ge \underline M,\\
\end{aligned}
\right\}.
\end{equation}

Joint limits are enforced explicitly in the refinement problem.

The set $\mathcal X_{\mathrm{term}}$ is determined by the robot model and the manipulation task. It is not an offline solution to the full navigation-- manipulation problem; rather, it specifies the class of base--arm preparation states from which the manipulation phase can be initiated. This remains compatible with online mapping: map updates may change the preferred route, the available clearance, or the feasibility of intermediate placements, but they do not generally change the target object or the terminal task requirement.

\subsection{Execution uncertainty and planning--control coupling}
\label{subsec:tracking_interface}

Nominal geometric feasibility is insufficient in confined underwater intervention if the realized base motion deviates significantly from the planned reference. Let $\mathbf r_t^{\mathrm{ref}}$ be the planned reduced base reference and let $\mathbf r_t^{\mathrm{cl}}$ be the realized closed-loop base motion. We assume that the terminal approach and station-keeping errors satisfy the bounded tracking interface $\mathbf r_t^{\mathrm{ref}}$
\begin{equation}
\label{eq:tracking_error_bound_rewritten}
\left\|
\mathbf r_t^{\mathrm{cl}}-\mathbf r_t^{\mathrm{ref}}
\right\|_{\Sigma_r^{-1}}
\le
\varepsilon_{\mathrm{trk}},
\qquad
 t\in[0,{N_{\mathrm{ref}}-1}],
\end{equation}
where $\|\mathbf e\|_{\Sigma_r^{-1}}=(\mathbf e^\top\Sigma_r^{-1}\mathbf e)^{1/2}$, $\Sigma_r\succ0$ sets the shape of the tracking-error tolerance, and $\varepsilon_{\mathrm{trk}}$ is a prescribed bound. Equation~\eqref{eq:tracking_error_bound_rewritten} is not intended as a complete stochastic reachability model. It is the planning--control interface through which expected execution error is translated into conservative geometric margins.

This coupling is essential in narrow passages: a clearance margin that is acceptable for a nominal path may be insufficient once closed-loop tracking error is taken into account. Therefore, both the selected corridor and the terminal operating state must retain safety margins compatible with the reach-and-hold performance of the controller.

\subsection{Formal problem statement}
\label{subsec:formal_problem_statement}

Given the current map $\mathcal M^\eta$, an initial configuration $\mathbf q_0$, a target frame ${}^{w}\!T_g$, and a UVMS model including geometry, kinematics, and actuation limits, the objective is to compute \(\mathcal S^\star=(\mathcal C^\star,\mathcal R^\star,\mathcal Q_a^\star,\mathbf q^\star,\pi_\theta)\)
where $\mathcal C^\star$ is a reduced base corridor, $\mathcal R^\star$ is a refined base trajectory, $\mathcal Q_a^\star$ is the corresponding arm trajectory, $\mathbf q^\star$ is the terminal operating state, and $\pi_\theta$ is the feedback policy used for execution.

We assume the following feasibility requirements holds:
\begin{enumerate}
    \item \emph{Access feasibility:} there exists a collision-safe reduced base
    corridor connecting the initial region to a neighborhood of the task region
    through the currently known confined free space.

    \item \emph{Placement feasibility:} along that corridor, there exists a
    refined base--arm trajectory whose terminal state satisfies \(\mathbf q^\star\in\mathcal X_{\mathrm{term}}.\)

    \item \emph{Execution feasibility:} there exists a feedback policy for the
    vehicle base that reaches and maintains the planned operating neighborhood
    with tracking error satisfying \eqref{eq:tracking_error_bound_rewritten}
    during terminal approach and station keeping.
\end{enumerate}

The resulting problem is therefore a coupled \emph{access--placement--execution} problem: determine a valid passage to the task region, determine a manipulation-ready operating state within that passage, and determine a controller that can realize and maintain that state under underwater uncertainty. The proposed method addresses these three requirements hierarchically.



\section{Proposed Approach}
\label{sec:method}

In this section, we introduce \emph{MANTA}, a hierarchical planning-and-control framework for confined underwater intervention. The framework addresses the access--placement--execution problem of Section~\ref{sec:problem_formulation} through three layers. The first layer performs passage-level reasoning in a conservative reduced base space. The second layer performs corridor-conditioned whole-body refinement, jointly optimizing the base motion and arm trajectory while preserving the selected passage class. The third layer executes the active reference using a learned base policy and monitors the remaining suffix against map updates.

For an initial map $\mathcal M^0$, a UVMS model $\mathcal U$, an initial configuration $\mathbf q_0$, and a target frame ${}^{w}\!T_g$, the nominal planning phase returns $(\mathcal P^0, \mathbf q^\star, \pi_\theta)$ where $\mathcal P^0=(\mathcal R^\star,\mathcal Q_a^\star)$ is a certified base--arm plan. During execution, the plan is map-versioned: it remains valid only while its remaining suffix is feasible under the current conservative map.

\begin{algorithm}[t]
\caption{MANTA nominal map-versioned planning}
\label{alg:manta_nominal_planning}
\begin{algorithmic}[1]
\Require map $\mathcal M^0$, model $\mathcal U$, initial state $\mathbf q_0$, target ${}^{w}\!T_g$
\Ensure plan $\mathcal P^0$, terminal state $\mathbf q^\star$, policy $\pi_\theta$
\State Set $\eta\gets0$ and build $d_{\mathrm c}^{0}$ \eqref{eq:conservative_distance_field}
\State Build the reduced lattice \eqref{eq:clearance_defs_compact}--\eqref{eq:base_validity}
\State Generate terminal candidates \eqref{eq:goal_hard_filter_refined}--\eqref{eq:goal_score_refined}
\State Extract topology-distinct corridors \eqref{eq:corridor_cost_refined}
\For{each corridor $\mathcal C^{(m)}$}
    \State Initialize base spline and arm knots
    \For{$t=0,\ldots,N_{\mathrm{alt}}-1$}
        \State Update the base spline \eqref{eq:alternating_bilevel_refined}
        \State Update the arm trajectory \eqref{eq:lower_objective_refined}
        \State Evaluate cross-section feasibility \eqref{eq:cross_section_terms_refined}
    \EndFor
    \State Score and certify the candidate \eqref{eq:final_corridor_score_refined}--\eqref{eq:certification_refined}
    \If{refinement invalidates the corridor}
        \State Retry with lattice-locked base refinement
    \EndIf
    \If{certified path violates terminal task tolerance}
        \State Repair the terminal segment and recheck \eqref{eq:certification_refined}
    \EndIf
\EndFor
\State Select the best certified candidate \eqref{eq:master_selection_refined}
\State Learn the execution policy $\pi_\theta$ with MC-PILCO
\State Store $\mathcal P^0$ and initialize the route-family graph $\mathcal G_{\mathrm R}^{0}$
\end{algorithmic}
\end{algorithm}

Algorithm~\ref{alg:manta_nominal_planning} summarizes the nominal planning pipeline. The remainder of this section details the reduced-space passage layer, the corridor-conditioned refinement layer, the online monitoring and repair logic, and the learned tracking policy. Implementation details such as cached lattices, local scene indices, and route databases are used for efficiency but do not change the mathematical structure of the method.

\subsection{Global passage reasoning in a conservative reduced space}
\label{sec:method_global}

Direct global search in the full UVMS configuration space would require discretizing both vehicle and arm variables. MANTA therefore separates route-family selection from whole-body refinement. The global layer searches over the reduced base state $\mathbf r=[x,y,z,\psi]^\top$, while the arm is held at the compact transit posture $\mathbf q_{a,\mathrm{tr}}$. 

At map version $\eta$, the environment is represented by the conservative signed distance field (SDF)
\begin{equation}
\label{eq:conservative_distance_field}
\begin{aligned}
d_{\mathrm c}^{\eta}(\mathbf x)
={}&
d_0^{\eta}(\mathbf x)
+
\mu^{\eta}(\mathbf x)
-
\beta_{\sigma}\sigma^{\eta}(\mathbf x),
\end{aligned}
\end{equation}
where $d_0^\eta$ is the nominal Euclidean signed distance field (ESDF) of the currently known occupied space, while $\mu^\eta$ and $\sigma^\eta$ are optional residual mean and uncertainty terms. In deterministic environments, the correction terms are omitted and $d_{\mathrm c}^{\eta}=d_0^\eta$. When the map version is fixed inside a planning
call, we write $d_{\mathrm c}$ for compactness.

The UVMS geometry is represented during screening and optimization by conservative spheres attached to selected vehicle, arm, and tool frames. If sphere $i$ has center $\mathbf c_i(\mathbf q)$ and radius $\rho_i$, the environment and self-clearance margins are
\begin{subequations}
\label{eq:clearance_defs_compact}
\begin{align}
h_i(\mathbf q)
&=
d_{\mathrm c}\!\left(\mathbf c_i(\mathbf q)\right)-\rho_i,
\label{eq:env_clearance_def}
\\
h_{ij}^{\mathrm{self}}(\mathbf q)
&=
\left\|\mathbf c_i(\mathbf q)-\mathbf c_j(\mathbf q)\right\|_2
-(\rho_i+\rho_j).
\label{eq:self_clearance_def}
\end{align}
\end{subequations}
These margins provide smooth conservative quantities during planning, whereas final acceptance is performed by the articulated collision checker. This use of distance-field clearance penalties follows the optimization-based motion-planning view in ~\cite{zucker2013chomp}.

With the arm fixed at $\mathbf q_{a,\mathrm{tr}}$, the reduced base-clearance
field is
\begin{equation}
\label{eq:phi_base}
\begin{aligned}
\phi_{\mathrm{base}}^{\eta}(\mathbf r)
=
\min_i
\Big[
 h_i(\mathbf q)
\Big] .
\end{aligned}
\end{equation}

The conservative reduced free space is
\begin{equation}
\label{eq:base_validity}
\mathcal L^{\eta}
=
\left\{
\mathbf r
\;\middle|\;
\phi_{\mathrm{base}}^{\eta}(\mathbf r)\ge\delta_b
\right\},
\end{equation}
where $\delta_b>0$ is the reduced-space safety margin.

A reduced-space safe state is not necessarily a useful terminal operating state for manipulation. Therefore, before running A*~\cite{hart1968formal}, i.e., a heuristic graph-search method that selects states using accumulated cost, MANTA constructs a terminal goal basin around the target.

For each terminal candidate
\begin{equation*}
\mathbf r_g=[(\mathbf p_g^{b})^\top,\psi_g]^\top,
\qquad
\mathbf p_g^b=[x_g,y_g,z_g]^\top,
\end{equation*}
we compute the reduced clearance $\phi_g=\phi_{\mathrm{base}}(\mathbf r_g)$,
the compact-posture tool residual $e_{\mathrm{tool},g}
=
\left\|
\mathbf p_e\!\left(\mathbf q\right)
-
\mathbf p_g
\right\|_2,$
the base--target distance $d_{\mathrm{bt},g}=\left\|\mathbf p_g-\mathbf p_g^b\right\|_2,$
and the target-facing yaw error
\begin{equation*}
e_{\psi,g}
=
\left|
\operatorname{wrap}\!\left(
\psi_g-
\operatorname{atan2}(p_{g,y}-y_g,p_{g,x}-x_g)
\right)
\right| .
\end{equation*}
Here $\mathbf p_e(\cdot)$ extracts the end-effector position from
$f_{\mathrm{fk}}(\cdot)$.

Because a candidate may be poor under the compact transit posture but still locally reachable after modest arm motion, MANTA also performs a cheap local reachability screen. Starting from $\mathbf q_{a,\mathrm{tr}}$, a bounded damped least-squares IK routine is run while holding the base at $\mathbf r_g$.
Let $\hat{\mathbf q}_{a,g}$ denote the resulting estimate. The local residual is $e_{\mathrm{reach},g}$ and the screened terminal error is $e_g=\min\left(e_{\mathrm{tool},g},e_{\mathrm{reach},g}\right).$

The same local IK routine provides the arm-deployment estimate $\Delta q_g
=
\left\|\hat{\mathbf q}_{a,g}-\mathbf q_{a,\mathrm{tr}}\right\|_2,$
and the coarse translational manipulability estimate
\begin{equation*}
m_g
=
\sqrt{\det\!\left(J_{g}J_{g}^\top+\varepsilon_J I\right)},
\end{equation*}
where $J_{g}$ is the arm-restricted translational end-effector Jacobian at 
$\hat{\mathbf q}_{a,g}$.

The strict terminal pool is
\begin{equation}
\label{eq:goal_hard_filter_refined}
\begin{aligned}
\mathcal G_{\mathrm{str}}
=
\Big\{
\mathbf r_g
\;\Big|\;&
\phi_g\ge\delta_b,
\quad
e_g\le\bar e_{\mathrm{tool}},
\\
& d_{\mathrm{bt},g}\le\bar d_{\mathrm{bt}},
\quad
e_{\psi,g}\le\bar e_\psi
\Big\} .
\end{aligned}
\end{equation}
A relaxed pool is retained when strict screening is too conservative:
\begin{equation}
\label{eq:goal_relaxed_filter_refined}
\begin{aligned}
\mathcal G_{\mathrm{rel}}
=
\Big\{
\mathbf r_g
\;\Big|\;&
\phi_g\ge\delta_b,
\quad
e_g\le\bar e_{\mathrm{rel}},
\\
& d_{\mathrm{bt},g}\le\bar d_{\mathrm{rel}},
\quad
e_{\psi,g}\le\bar e_{\psi,\mathrm{rel}}
\Big\}
\setminus
\mathcal G_{\mathrm{str}} .
\end{aligned}
\end{equation}
The relaxed thresholds satisfy
$\bar e_{\mathrm{rel}}\ge\bar e_{\mathrm{tool}}$,
$\bar d_{\mathrm{rel}}\ge\bar d_{\mathrm{bt}}$, and
$\bar e_{\psi,\mathrm{rel}}\ge\bar e_\psi$.

The retained candidates are ranked by
\begin{equation}
\label{eq:goal_score_refined}
\begin{aligned}
\sigma_g
={}&
w_e e_g
+
w_d d_{\mathrm{bt},g}
+
w_\psi e_{\psi,g}
-
w_\phi \phi_g
\\
&-
w_m m_g
+
w_q\Delta q_g .
\end{aligned}
\end{equation}
This score favors small terminal error, short standoff, target-facing yaw, large clearance, high local manipulability, and limited arm deployment. A candidate is passed to corridor extraction only if the validated terminal residual and conservative clearance satisfy the required task tolerances as defined in \eqref{eq:terminal_set_rewritten}.

For each selected terminal state, MANTA extracts one or more reduced corridors in $\mathcal L^{\eta}$ using A* search on the four-dimensional lattice. For neighboring states $\mathbf r_i=[\mathbf p_i^\top,\psi_i]^\top$ and
$\mathbf r_j=[\mathbf p_j^\top,\psi_j]^\top$, the transition cost is
\begin{equation}
\label{eq:corridor_cost_refined}
\begin{aligned}
c(\mathbf r_i,\mathbf r_j)
={}&
w_\ell
\left(
\left\|\mathbf p_j-\mathbf p_i\right\|_2
+
0.2\,\left|\operatorname{wrap}(\psi_j-\psi_i)\right|
\right)
\\
&+
\frac{w_{\mathrm{clr}}}
{\min\{\phi_{\mathrm{base}}(\mathbf r_i),
       \phi_{\mathrm{base}}(\mathbf r_j)\}+\varepsilon}
\\
&+
w_{\mathrm{risk}}\,
\sigma\!\left(\frac{\mathbf p_i+\mathbf p_j}{2}\right)
+
\Pi(\mathbf r_j) .
\end{aligned}
\end{equation}
The first term penalizes translation and yaw variation, the second discourages low-clearance transitions, the third accounts for optional map risk, and $\Pi(\cdot)$ is a revisit penalty used to extract corridor-distinct alternatives. After each corridor extraction, a fixed penalty is assigned to its lattice states, producing a small route family rather than repeated versions of the same shortest path. This mechanism is related to search-based planning over distinct homotopy classes, although MANTA uses a pragmatic revisit-penalty heuristic rather than an explicit homotopy invariant~\cite{bhattacharya2010homotopy}.

\subsection{Corridor-conditioned local refinement and certification}
\label{sec:method_bilevel}

The global layer returns a topologically meaningful corridor, but not a complete UVMS trajectory. The refinement layer converts each candidate corridor into a continuous base path and an articulated arm trajectory. The corridor is used as a topological guide: the optimizer may smooth and shift the path inside a local tube, but it should not change the selected passage class.

Let $\Gamma^{(m)}$ denote the polyline centerline of corridor $\mathcal C^{(m)}$,
with discrete states $\mathbf g_j^{(m)}=[(\mathbf p_j^\Gamma)^\top,\psi_j^\Gamma]^\top.$

Each vertex is assigned a normalized arclength station
$\bar u_j=s_j/L^{(m)}\in[0,1],$ where $s_j$ is the cumulative arclength up to vertex $j$ and $L^{(m)}$ is the total corridor length. The continuous base trajectory is represented by
a cubic open B-spline
\begin{equation*}
\label{eq:base_spline_parameterization}
\mathbf r(u)
=
\sum_{i=0}^{n_c^{(m)}-1}
N_{i,3}(u;\mathcal U^{(m)})\,\boldsymbol\theta_i,
\qquad
u\in[0,1],
\end{equation*}
where $\boldsymbol\theta_i\in\mathbb R^4$ are position--yaw control states.

The number and placement of control states are adapted to the selected corridor. MANTA first extracts an anchor set $\mathcal A^{(m)}$ containing the start, goal, strong turning points, vertical reversal points, and low-clearance local minima.
The number of control states is
\begin{equation*}
\label{eq:adaptive_control_count}
\begin{aligned}
\bar n_c^{(m)}
&=
\max\Bigg\{
 n_c^{\min},
 \left\lceil\frac{L^{(m)}}{\Delta_c}\right\rceil+1,
 |\mathcal A^{(m)}|+2
\Bigg\},
\\
n_c^{(m)}
&=
\min\left\{n_c^{\max},\bar n_c^{(m)}\right\} .
\end{aligned}
\end{equation*}
Control stations are formed from anchor stations, shoulder points around non-terminal anchors, and additional mid-gap stations until the control budget is reached. This gives the spline extra support near gates, vertical reversals, and sharp bends while retaining a compact representation in open regions.

Let
\(\mathcal R(\Theta)=\{\mathbf r_k(\Theta)\}_{k=0}^{N}\) denote the sampled base path induced by the spline control variables \(\Theta\). For a fixed corridor, the upper-level update optimizes the base spline through the objective
\begin{equation*}
\label{eq:upper_objective_refined}
\begin{aligned}
J_{\mathrm{up}}
={}&
w_\ell J_\ell
+
w_{\mathrm{tube}}J_{\mathrm{tube}}
+
J_{\mathrm{anc}}
+
w_{\mathrm{clr}}J_{\mathrm{clr}}
\\
&+
w_{\mathrm{align}}J_{\mathrm{align}}
+
w_{\mathrm{sm}}J_{\mathrm{sm}}
+
\lambda_{\mathrm{low}}J_{\mathrm{low}}^{\mathrm{sur}} .
\end{aligned}
\end{equation*}
The first group of terms shapes the nominal base motion: \(J_\ell\) penalizes path length, \(J_{\mathrm{sm}}\) regularizes curvature and yaw variation, and \(J_{\mathrm{align}}\) keeps the vehicle heading consistent with the selected corridor when a reliable yaw reference exists. The corridor terms \(J_{\mathrm{tube}}\) and \(J_{\mathrm{anc}}\) preserve the route selected by the global planner. In particular, \(J_{\mathrm{tube}}\) penalizes deviation from the corridor at the same normalized arclength station, while \(J_{\mathrm{anc}}\) keeps critical events such as turns, vertical reversals, and low-clearance gates from being smoothed away. The clearance term \(J_{\mathrm{clr}}\) biases the path toward safer regions inside the corridor neighborhood. Finally, \(J_{\mathrm{low}}^{\mathrm{sur}}\) transfers the dominant arm-dependent effects to the base update, including environment clearance, cross-section occupancy, and terminal task consistency, without differentiating through a fully solved lower-level arm optimization.

Throughout this subsection we use the smooth penalty primitives
\[
\operatorname{smax}_{\kappa}\!\big(\{a_i\}\big)
=
\kappa^{-1}\log\!\Big(\sum_i e^{\kappa a_i}\Big),
\]
and
\[
\operatorname{smin}_{\tau}\!\big(\{a_i\}\big)
=
-\tau\log\!\Big(\sum_i e^{-a_i/\tau}\Big),
\]
where $\kappa>0$ and $\tau>0$ are smoothing parameters.
These operators convert hard geometric constraints into differentiable penalties while preserving conservative behavior near the boundary of feasibility.

The upper-level clearance is tightened across alternating iterations:
\begin{equation*}
\label{eq:anchored_clearance_refined}
\bar\phi_k
=
\begin{cases}
\phi_{\mathrm{base}}(\mathbf r_k),
& t=0,
\\[1mm]
\operatorname{smin}_{\tau_{\mathrm{clr}}}
\left\{
 h_i(\mathbf q_k)
\right\}_{i},
& t\ge1 .
\end{cases}
\end{equation*}
The first upper update uses the conservative reduced base model, while later updates use the current articulated arm anchor.

For a fixed base path, the lower level optimizes the arm at a sparse set of critical knots. Let $\mathcal K=\{\kappa_1,\ldots,\kappa_M\}$ be the active arm-knot indices. The set contains endpoints, terminal samples, low-clearance samples, narrow cross-section samples, neighboring samples around critical events, and support knots. The arm at all path samples is obtained by interpolation between optimized knots $\mathbf q_{a,k}=\mathcal I_k(\mathbf z_1,\ldots,\mathbf z_M)$ where $\mathbf z_j$ is the active-joint vector at knot $\kappa_j$.

The lower-level objective is
\begin{equation}
\label{eq:lower_objective_refined}
\begin{aligned}
J_{\mathrm{low}}
={}&
\sum_{k\in\mathcal E}\Phi_{\mathrm{col},k}
+
w_{\mathrm{stow}}
\sum_{k\in\mathcal E_{\mathrm{tr}}}
\left\|\mathbf q_{a,k}-\mathbf q_{a,\mathrm{tr}}\right\|_2^2
\\
&+
w_{\mathrm{sm}}
\sum_{j=1}^{M-1}
\frac{\left\|\mathbf z_{j+1}-\mathbf z_j\right\|_2^2}
     {\kappa_{j+1}-\kappa_j}
+
w_{\mathrm{task}}\Phi_{\mathrm{task}}(\mathbf q_N)
\\
&-
w_{\mathrm{man}}M_{\mathrm{dir}}(\mathbf q_N) .
\end{aligned}
\end{equation}
Here $\mathcal E$ is the evaluation set formed by optimized knots, local
neighbors, and midpoints between consecutive knots. The set $\mathcal E_{\mathrm{tr}}$
denotes the early transit portion where compact-posture regularization is active.

The aggregate collision and feasibility penalty is
\begin{equation*}
\label{eq:lower_collision_aggregate}
\begin{aligned}
\Phi_{\mathrm{col},k}
={}&
w_{\mathrm{env}}\Phi_{\mathrm{env},k}
+
w_{\mathrm{self}}\Phi_{\mathrm{self},k}
\\
&+
w_{\mathrm{lim}}\Phi_{\mathrm{lim},k}
+
w_{\mathrm{cross}}\Phi_{\mathrm{cross},k} .
\end{aligned}
\end{equation*}
The constituent terms include
\begin{subequations}
\label{eq:lower_terms_refined}
\begin{align}
\Phi_{\mathrm{env},k}
&=
\sum_i
\frac{1}{2}
\left[
\operatorname{sp}\!\left(m_{\mathrm{env}}-h_i(\mathbf q_k)\right)
\right]^2,
\\
\Phi_{\mathrm{self},k}
&=
\sum_{(i,j)\in\mathcal P_{\mathrm{self}}}
\frac{1}{2}
\left[
\operatorname{sp}\!\left(
 m_{\mathrm{self}}-h_{ij}^{\mathrm{self}}(\mathbf q_k)
\right)
\right]^2,
\\
\Phi_{\mathrm{task}}(\mathbf q_N)
&=
\left\|\mathbf p_e(\mathbf q_N)-\mathbf p_g\right\|_2^2 .
\end{align}
\end{subequations}
The joint-limit term $\Phi_{\mathrm{lim},k}$ is implemented as a smooth barrier
around the active joint bounds.

Terminal dexterity is evaluated directionally when task directions are specified.
For motion directions $\mathbf v_r$ and force directions $\mathbf f_s$,
\begin{subequations}
\label{eq:directional_manip_refined}
\begin{align*}
\alpha_r(\mathbf q_N)
&=
\left(
\mathbf v_r^\top(JJ^\top)^{-1}\mathbf v_r
\right)^{-1/2},
\\
\beta_s(\mathbf q_N)
&=
\left(
\mathbf f_s^\top(JJ^\top)\mathbf f_s
\right)^{-1/2} .
\end{align*}
\end{subequations}
The directional manipulability score $M_{\mathrm{dir}}$ is a weighted sum of these quantities. For position-only targets, the planner falls back to translational manipulability of the active positioning joints in the remaining target direction.

To prevent the articulated body from expanding laterally beyond the available aperture, the lower objective includes a corridor cross-section penalty. At sample $k$, let $\boldsymbol\gamma_k$ be the associated corridor centerline point, $\boldsymbol\tau_k$ the local tangent, and $\{\boldsymbol\tau_k,\mathbf n_k,\mathbf b_k\}$ the corresponding orthonormal frame. Let $(a_k,b_k)$ be conservative free-space semiaxes along $\mathbf n_k$ and $\mathbf b_k$. The projected robot span is
\begin{subequations}
\label{eq:cross_section_terms_refined}
\begin{align}
\rho_{n,k}
&=
\operatorname{smax}_{\kappa}
\left\{
\left|
\mathbf n_k^\top(\mathbf c_i(\mathbf q_k)-\boldsymbol\gamma_k)
\right|+\rho_i
\right\}_{i},
\\
\rho_{b,k}
&=
\operatorname{smax}_{\kappa}
\left\{
\left|
\mathbf b_k^\top(\mathbf c_i(\mathbf q_k)-\boldsymbol\gamma_k)
\right|+\rho_i
\right\}_{i},
\\
\ell_k
&=
\frac{\rho_{n,k}^2}{(a_k-m_{\mathrm{cross}})^2}
+
\frac{\rho_{b,k}^2}{(b_k-m_{\mathrm{cross}})^2},
\\
\Phi_{\mathrm{cross},k}
&=
\frac{1}{2}
\left[
\operatorname{sp}(\ell_k-1)
\right]^2 .
\end{align}
\end{subequations}

\begin{figure*}[t]
    \centering
    \captionsetup[subfigure]{font=small}

    \begin{subfigure}[t]{0.51\textwidth}
        \centering
        \includegraphics[width=\linewidth,height=0.20\textheight,keepaspectratio]{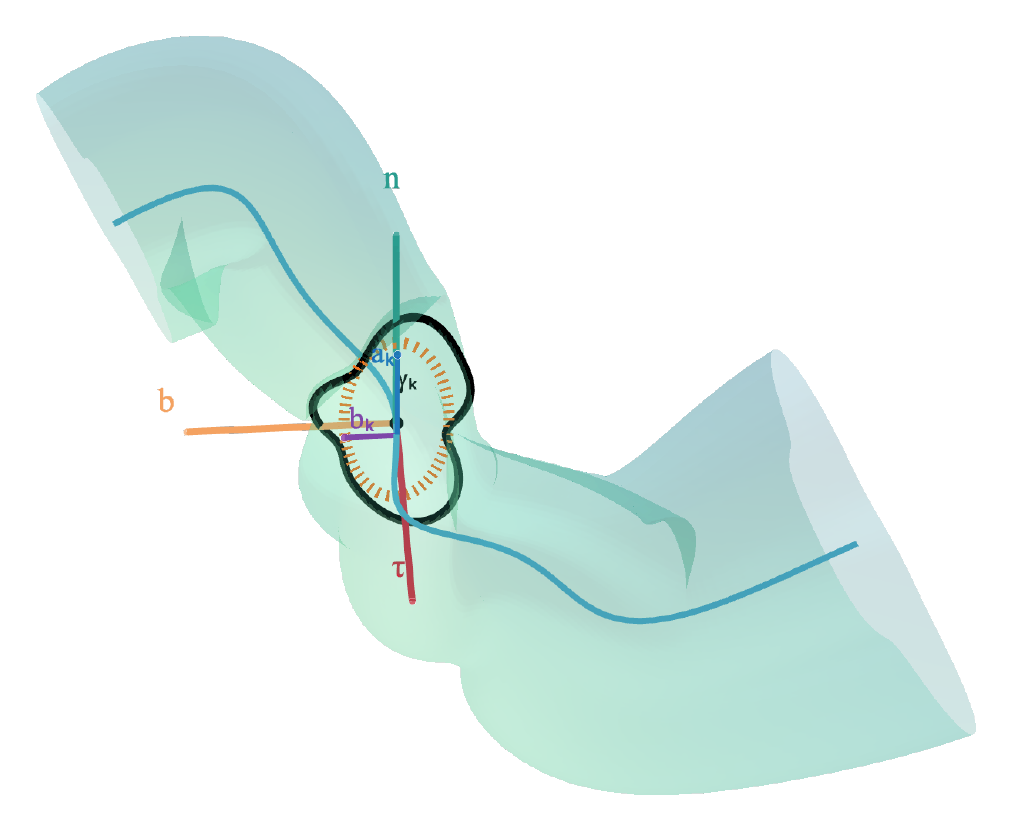}
        \caption{Local 3D corridor slice and cross-section frame.}
        \label{fig:cross_section_wav_3D}
    \end{subfigure}
    \hfill
    \begin{subfigure}[t]{0.48\textwidth}
        \centering
        \includegraphics[width=\linewidth,height=0.20\textheight,keepaspectratio]{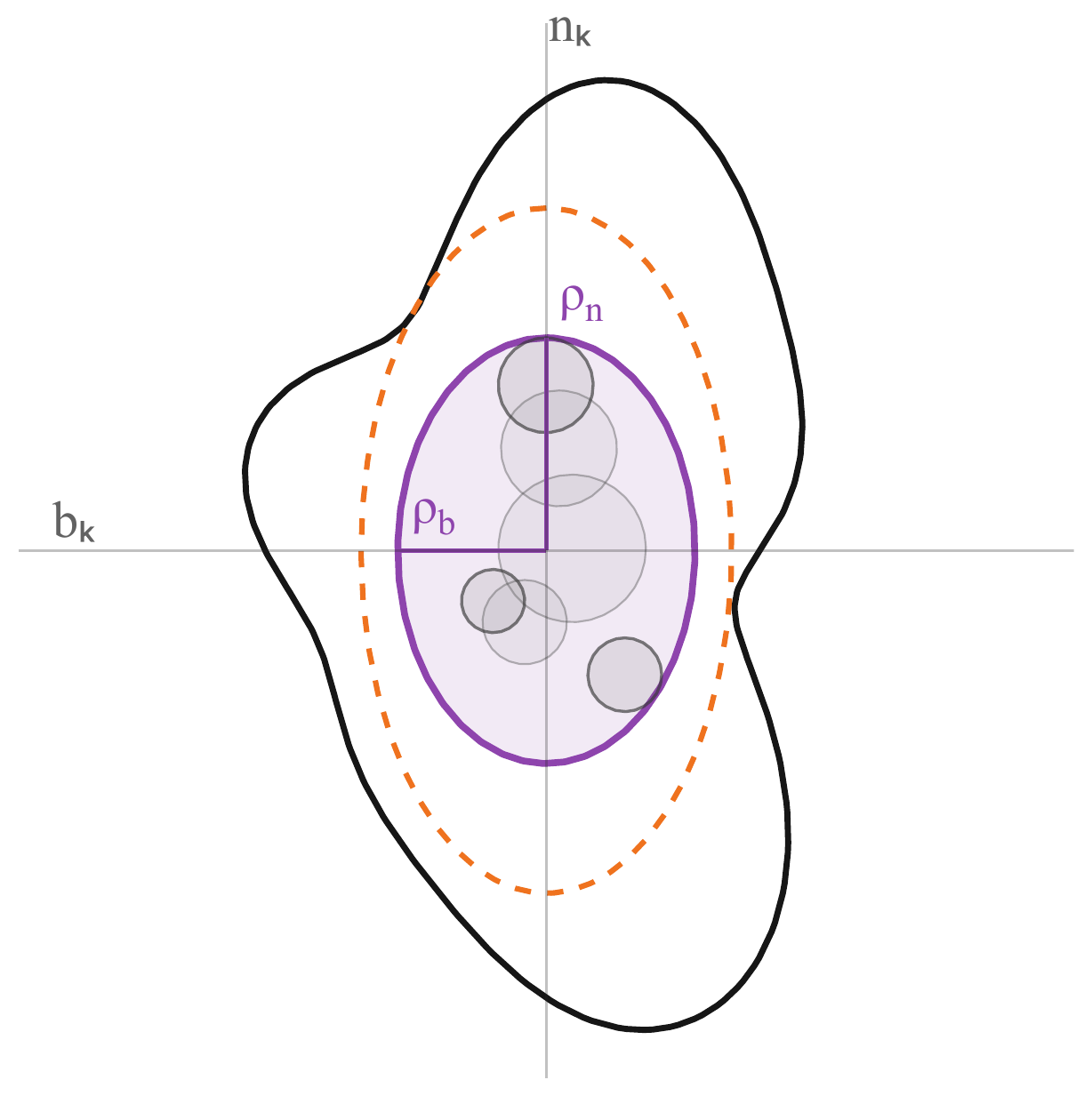}
        \caption{Cross-section-plane projection used to evaluate $\Phi_{\mathrm{cross},k}$.}
        \label{fig:cross_section_wavy_2D}
    \end{subfigure}

    \caption{Corridor cross-section penalty at sample $k$.
        At $\boldsymbol{\gamma}_k$, the local frame
        $\{\boldsymbol{\tau}_k,\mathbf n_k,\mathbf b_k\}$ defines the normal plane.
        The true corridor boundary is shown in black and is conservatively approximated
        by the \emph{orange dashed ellipse} with semiaxes $(a_k,b_k)$, while the
        projected UVMS sphere model is represented by the \emph{purple solid ellipse}
        through $\rho_{n,k}$ and $\rho_{b,k}$.}
    \label{fig:cross_section_penalty_panel}
\end{figure*}

This term penalizes the projected spread of the whole articulated body relative to the local aperture, and is therefore stronger than pointwise clearance alone in tube-like and cave-like passages.

The upper and lower levels are solved in alternation. At iteration $t$,
\begin{subequations}
\label{eq:alternating_bilevel_refined}
\begin{align}
\Theta^{(t+1,m)}
&=
\arg\min_{\Theta}
J_{\mathrm{up}}(\Theta,\bar{\mathcal Q}_a^{(t,m)}),
\\
\mathcal Q_a^{(t+1,m)}
&=
\arg\min_{\mathcal Q_a}
J_{\mathrm{low}}\!\left(
\mathcal Q_a;
\mathcal R(\Theta^{(t+1,m)}),
\mathcal C^{(m)}
\right).
\end{align}
\end{subequations}
Since the spline basis and sampling grid are fixed after corridor selection, the
sampled path remains affine in the control states. If
$\mathcal R=B\Theta$, $\dot{\mathcal R}=B'\Theta$, and
$\ddot{\mathcal R}=B''\Theta$, then
\begin{equation*}
\label{eq:control_chain_rule_refined}
\nabla_\Theta J
=
B^\top\nabla_{\mathcal R}J
+
(B')^\top\nabla_{\dot{\mathcal R}}J
+
(B'')^\top\nabla_{\ddot{\mathcal R}}J .
\end{equation*}

In tight S-bends or keyholes, continuous smoothing may occasionally turn a valid lattice corridor into an invalid base footprint. MANTA therefore includes a lattice-locked fallback. If continuous refinement fails, the planner keeps the base and yaw exactly on the valid discrete corridor, optionally downsamples very long corridors, and solves only the lower arm problem.

After refinement, each corridor receives the score
\begin{equation}
\label{eq:final_corridor_score_refined}
\begin{aligned}
S^{(m)}
={}&
J_{\mathrm{up}}(\Theta_\star^{(m)})
+
\lambda_{\mathrm{sel}}
J_{\mathrm{low}}\!\left(
\mathcal Q_{a,\star}^{(m)};
\mathcal R_\star^{(m)},
\mathcal C^{(m)}
\right) .
\end{aligned}
\end{equation}
where $\lambda_{\mathrm{sel}}$ is the selection weight used to compare refined candidates. The final candidate is selected only among certified trajectories:
\begin{equation}
\label{eq:master_selection_refined}
\begin{aligned}
m^\star
\in
\arg\min_m S^{(m)}
\quad
\mathrm{s.t.}\quad
\mathrm{Cert}^{\eta}\!\left(
\mathcal R_\star^{(m)},
\mathcal Q_{a,\star}^{(m)}
\right)=1 .
\end{aligned}
\end{equation}
Certification is performed with the exact Pinocchio/HPP-FCL collision model~\cite{carpentier2019pinocchio} and
the current map version:
\begin{equation}
\label{eq:certification_refined}
\begin{aligned}
\mathrm{Cert}^{\eta}(\mathcal R,\mathcal Q_a)=1
\iff{}&
\mathbf q(\mathbf r_k,\mathbf q_{a,k})
\in
\mathcal Q_{\mathrm{free}}^{\eta},
\\
&k=0,\ldots,N .
\end{aligned}
\end{equation}

Finally, MANTA applies a local terminal repair when a trajectory is geometrically certified but its terminal tool error remains above the task tolerance. The repair optimizes only the last $L$ samples, using active arm joints and a bounded final yaw correction that is ramped smoothly over the segment. The repair is accepted only if it improves the terminal error, preserves certification, and does not introduce excessive arm motion.

\subsection{Online feasibility monitoring and replanning}
\label{sec:method_online_replanning}

\begin{algorithm}[t]
\caption{Online feasibility monitoring and replanning}
\label{alg:online_replanning}
\begin{algorithmic}[1]
\Require active plan $\mathcal P^\eta$, route graph $\mathcal G_{\mathrm R}^{\eta}$, policy $\pi_\theta$, map $\mathcal M^\eta$
\Ensure updated plan $\mathcal P^\eta$ or safe-hold report
\While{the task is not completed}
    \State Execute $\pi_\theta$ and update the closest plan index $k_{\mathrm c}$
    \If{a map update is received}
        \State Update $\eta$, $d_{\mathrm c}^{\eta}$, and $\mathcal Z^\eta$
        \If{\eqref{eq:update_intersection_filter} holds}
            \State Retain the active plan
        \ElsIf{$\chi^\eta(k_{\mathrm c})\ge\delta_{\mathrm{exec}}$ by \eqref{eq:online_suffix_feasibility}}
            \State Retain the active plan
        \Else
            \State Hold at $\bar{\mathbf r}$ and recover the arm using \eqref{eq:fixed_base_arm_recovery}
            \State Update $\mathcal G_{\mathrm R}^{\eta}$ using \eqref{eq:route_edge_status_update}
            \State Search for direct or branch repair by \eqref{eq:direct_repair_filter}--\eqref{eq:branch_aware_repair}
            \If{a repair is refined and certified}
                \State Replace active plan by the certified repair
            \Else
                \State Return safe-hold report
            \EndIf
        \EndIf
    \EndIf
\EndWhile
\end{algorithmic}
\end{algorithm}

The nominal planner certifies a base--arm trajectory with respect to the conservative map available at planning time. In partially known environments, new obstacle or risk information may become available during execution. MANTA therefore triggers replanning only when a map update invalidates the remaining executable suffix of the active plan.

The logic is related to incremental replanning methods~\cite{koenig2002dstar} that reuse previous search information after local map changes, but here the update is coupled to suffix validation, arm recovery, and corridor repair for the UVMS setting.

Let the active certified plan at map version $\eta$ be $\mathcal P^\eta=\{\mathbf q_k^\eta\}_{k=0}^{N},$ and let $k_{\mathrm c}$ be the current closest index on the plan. The remaining
suffix is $\mathcal P_{k_{\mathrm c}:}^{\eta}
=
\{\mathbf q_k^\eta\}_{k=k_{\mathrm c}}^{N} .$ When a map update is received, the conservative field $d_{\mathrm c}^{\eta}$ is rebuilt and the update is summarized by the affected zone
\begin{equation}
\label{eq:affected_region}
\mathcal Z^\eta
=
\left\{
\mathbf x\in\mathbb R^3
\;\middle|\;
\mathbf z_{\min}^{\eta}
\preceq
\mathbf x
\preceq
\mathbf z_{\max}^{\eta}
\right\} .
\end{equation}

Before checking all remaining trajectory samples, the planner tests whether the affected zone intersects a conservative swept envelope of the suffix. Let $\mathcal I(k_{\mathrm c})$ be the suffix indices selected for online checking and let $\rho_Z>0$ be an additional padding. The suffix envelope is
\begin{equation*}
\label{eq:suffix_envelope}
\mathcal E_{k_{\mathrm c}}^\eta
=
\bigcup_{k\in\mathcal I(k_{\mathrm c})}
\bigcup_i
\mathbb B\!\left(
\mathbf c_i(\mathbf q_k^\eta),
\rho_i+\rho_Z
\right).
\end{equation*}
If this envelope does not intersect the affected map region, the suffix is
retained:
\begin{equation}
\label{eq:update_intersection_filter}
\mathcal Z^\eta\cap\mathcal E_{k_{\mathrm c}}^\eta=\emptyset
\quad\Longrightarrow\quad
\mathcal P_{k_{\mathrm c}:}^{\eta-1}
\text{ is retained.}
\end{equation}

If the affected region intersects the suffix envelope, the remaining trajectory is explicitly rechecked. The checked set $\mathcal S(k_{\mathrm c})$ contains selected suffix configurations and interpolated configurations along consecutive segments. The online suffix clearance is
\begin{equation}
\label{eq:online_suffix_feasibility}
\chi^\eta(k_{\mathrm c})
=
\min_{\mathbf q\in\mathcal S(k_{\mathrm c})}
\;
\min_i
\left[
 d_{\mathrm c}^{\eta}\!\left(\mathbf c_i(\mathbf q)\right)
 -\rho_i
\right] .
\end{equation}
The active suffix remains executable if $\chi^\eta(k_{\mathrm c})\ge\delta_{\mathrm{exec}}$. Otherwise, the vehicle switches to hold-and-repair mode.

If the arm is not already close to the transit posture, MANTA first solves a fixed-base recovery problem. Let $\bar{\mathbf r}$ be the held base pose, $\mathbf q_{a,\mathrm{cur}}$ the current arm state, and $\mathbf q_{a,\mathrm{tr}}$ the transit posture. The recovery trajectory $\mathcal Q_a^{\mathrm{rec}}=\{\mathbf q_{a,s}^{\mathrm{rec}}\}_{s=0}^{N_r}$ is obtained from
\begin{equation}
\label{eq:fixed_base_arm_recovery}
\begin{aligned}
\mathcal Q_a^{\mathrm{rec}}
=
\arg\min_{\mathcal Q_a}\quad
&J_r(\mathcal Q_a;\bar{\mathbf r})
\\
\mathrm{s.t.}\quad
&\mathbf q_{a,0}=\mathbf q_{a,\mathrm{cur}},
\\
&\mathbf q_a^{\min}\preceq\mathbf q_{a,s}\preceq\mathbf q_a^{\max},
\quad s=0,\ldots,N_r .
\end{aligned}
\end{equation}
The recovery objective is
\begin{equation*}
\label{eq:recovery_objective}
\begin{aligned}
J_r
={}&
\sum_{s=0}^{N_r}\Phi_r(\bar{\mathbf r},\mathbf q_{a,s})
+ w_{\mathrm{sm}}^r
\sum_{s=0}^{N_r-1}
\left\|\mathbf q_{a,s+1}-\mathbf q_{a,s}\right\|_2^2
\\
&+
w_{\mathrm{tr}}^r
\left\|\mathbf q_{a,N_r}-\mathbf q_{a,\mathrm{tr}}\right\|_2^2
+
w_{\mathrm{post}}^r
\sum_{s=0}^{N_r}
\left\|\mathbf q_{a,s}-\mathbf q_{a,\mathrm{tr}}\right\|_2^2 .
\end{aligned}
\end{equation*}
Here $N_{\mathrm r}$ is the number of recovery waypoints, $\Phi_{\mathrm{env}}$ is the same environment penalty used in \eqref{eq:lower_terms_refined}, and the superscript ${\mathrm r}$ denotes recovery-specific weights. The smoothness term regularizes the recovery motion, while the terminal and posture terms bias the final arm state toward the transit posture. The recovery is accepted only if the terminal arm error is below tolerance and the minimum conservative clearance along the recovery path remains above
$\delta_{\mathrm{exec}}$.

After the robot reaches a safe route-repair configuration, the planner performs Layer-1 repair under the updated map. To avoid treating the initial corridor as a disposable one-shot path, the framework stores the family of Layer-1 corridors in a route-family graph
$
\mathcal G_{\mathrm R}^{\eta}
=
(\mathcal V^{\eta},\mathcal E^{\eta}).
$
Each node $v\in\mathcal V^\eta$ corresponds to a salient reduced-space state, such as a start state, terminal goal state, low-clearance gate, turn, or previously identified branch point. Each edge $e\in\mathcal E^\eta$ stores a corridor segment $\mathcal C_e$, its map version, an axis-aligned swept bounding box $\mathcal A_e$, and a status
$
s_e^\eta\in\{\mathrm{valid},\mathrm{unknown},\mathrm{invalid}\}.
$
When an affected map region $\mathcal Z^\eta$ is produced by a map update, only graph edges whose bounding boxes intersect the affected map region are re-evaluated:
\begin{equation}
\label{eq:route_edge_status_update}
s_e^\eta
=
\begin{cases}
s_e^{\eta-1},
&
\mathcal A_e\cap\mathcal Z^\eta=\emptyset,
\\[1mm]
\mathrm{invalid},
&
\mathcal A_e\cap\mathcal Z^\eta\neq\emptyset
\;\land\;
\displaystyle
\min_{\mathbf r\in\mathcal C_e}
\phi_{\mathrm{base}}^\eta(\mathbf r)
<\delta_{\mathrm g},
\\[3mm]
\mathrm{unknown},
&
\mathcal A_e\cap\mathcal Z^\eta\neq\emptyset
\;\land\;
\displaystyle
\max_{\mathbf r\in\mathcal C_e}
\sigma^\eta(\mathbf p(\mathbf r))
\ge \bar\sigma_{\mathrm g},
\\[3mm]
\mathrm{valid},
&
\text{otherwise}.
\end{cases}
\end{equation}
Here $\phi_{\mathrm{base}}^\eta$ is the reduced-space clearance field computed from $d_{\mathrm c}^{\eta}$ as in \eqref{eq:phi_base}, $\delta_{\mathrm g}>0$ is the graph-level clearance margin, $\sigma^\eta$ is the map-risk indicator used in \eqref{eq:corridor_cost_refined}, and $\bar\sigma_{\mathrm g}$ is the threshold above which an edge is treated as uncertain rather than fully validated.

The repair search first attempts a direct route from the current held state
$
\mathbf r_{\mathrm c}
$
to the selected terminal goal state
$
\mathbf r_g^\star.
$
However, a route that immediately moves backward through the already traversed passage is not classified as a direct repair. Let
$
\mathbf e_{\mathrm{cg}}
=
(\mathbf p_g^\star-\mathbf p_{\mathrm c})/
\|\mathbf p_g^\star-\mathbf p_{\mathrm c}\|_2
$
be the current-to-goal direction. A direct candidate corridor $\mathcal C$ is accepted as forward-progressing only if
\begin{equation}
\label{eq:direct_repair_filter}
\min_{\mathbf r\in\mathcal C}
\left(
\mathbf p(\mathbf r)-\mathbf p_{\mathrm c}
\right)^\top
\mathbf e_{\mathrm{cg}}
\ge
-\delta_{\mathrm{back}},
\end{equation}
where $\delta_{\mathrm{back}}\ge 0$ is a small tolerance that permits local maneuvering but rejects corridors whose main effect is to return to a previous branch.

If no direct forward corridor can be certified, the route-family graph is used to search through previous branch or gate states. Let $\mathcal B(k_c)$ be the set of graph nodes whose stored progress is at least $\Delta_u>0$ behind the current progress. For each candidate branch node $b\in\mathcal B(k_c)$,
\begin{equation}
\label{eq:branch_aware_repair}
\begin{aligned}
\mathcal C_b^{\mathrm{rep}}
&=
\mathcal C_{c\rightarrow b}
\oplus
\mathcal C_{b\rightarrow g},
\\
\mathcal C_{c\rightarrow b}
&\in
\mathrm{Extract}\!\left(
\mathbf r_c,\mathbf r_b;\mathcal M^\eta
\right),
\\
\mathcal C_{b\rightarrow g}
&\in
\mathrm{Extract}\!\left(
\mathbf r_b,\mathbf r_g^\star;\mathcal M^\eta
\right).
\end{aligned}
\end{equation}
The operator $\oplus$ concatenates corridors with duplicate junction states removed. A branch repair is accepted at Layer 1 only if both subcorridors are connected in the updated conservative lattice and pass the swept reduced-body clearance test. The resulting repaired corridor is then passed to the same Layer-2 refinement and exact certification machinery used by nominal planning.

\subsection{Learning-based trajectory tracking policy via MC-PILCO}
\label{subsec:mcpilco_tracking}

The planning and repair layers return a collision-certified base--arm motion and a reduced base reference that must be realized during execution. Since geometric feasibility is meaningful only up to the tracking tolerance in \eqref{eq:tracking_error_bound_rewritten}, the execution layer learns a feedback policy that tracks the active reference while compensating for hydrodynamic mismatch, actuator effects, environmental disturbances, and base--arm coupling.

We use MC-PILCO as a data-efficient MBRL layer. At each trial, a probabilistic dynamics model is fitted from closed-loop data, the policy is optimized through Monte Carlo rollouts of the learned model, and the updated controller is executed to collect additional data.

Let the reduced pose reference generated by the planning layer be $\mathcal R_{\mathrm{ref}}
=
\{\mathbf r_t^{\mathrm{ref}}\}_{t=0}^{N_{\mathrm{ref}}-1}.$
For tracking, this pose reference is augmented with desired body-frame velocities,
\[
\bar{\mathbf x}^{\mathrm{ref}}_t
=
\left[
(\mathbf r^{\mathrm{ref}}_t)^\top,
(\boldsymbol{\nu}^{\mathrm{ref}}_t)^\top
\right]^\top,
\quad
\boldsymbol{\nu}^{\mathrm{ref}}_t
=
[u^{\mathrm{ref}}_t,
v^{\mathrm{ref}}_t,
w^{\mathrm{ref}}_t,
r^{\mathrm{ref}}_t]^\top .
\]
The learning state is $\mathbf x_t
=
\left[
\mathbf r_t^\top,
\boldsymbol\nu_t^\top,
 t
\right]^\top$
where $t$ is the reference phase index.

\subsubsection{Probabilistic dynamics model}
\label{subsubsec:mcpilco_dynamics_model}

The GP model predicts one-step velocity increments, 
\begin{equation*}
\Delta\boldsymbol\nu_t
=
[\Delta u_t,\Delta v_t,\Delta w_t,\Delta r_t]^\top .
\end{equation*}
Each component is modeled by an independent Gaussian process,
\begin{equation*}
\label{eq:mcpilco_delta_model}
\Delta\nu_{i,t}
=
f_i(\mathbf z_t)+\epsilon_i,
\qquad
\epsilon_i\sim\mathcal N(0,\sigma_i^2),
\end{equation*}
with regressor
\begin{equation*}
\label{eq:mcpilco_gp_regressor}
\mathbf z_t
=
[\mathbf s_t^\top,\sin\psi_t,\cos\psi_t,\mathbf a_t^\top]^\top .
\end{equation*}
Here $\mathbf a_t$ is the thruster command and $\mathbf s_t$ contains the selected non-angular local state components. Absolute position may be omitted from $\mathbf s_t$ to obtain a translation-invariant local dynamics representation.

Given a sampled GP prediction,
\begin{equation*}
\boldsymbol\nu_{t+1}=\boldsymbol\nu_t+\Delta\boldsymbol\nu_t .
\end{equation*}
The reduced pose is reconstructed through known kinematics:
\begin{equation*}
\label{eq:mcpilco_state_propagation}
\begin{aligned}
\psi_{t+1}
&=
\psi_t+\frac{T_s}{2}(r_t+r_{t+1}),
\\
\mathbf p_{t+1}
&=
\mathbf p_t
+
T_s R_z(\bar\psi_t)
\frac{\mathbf v_t+\mathbf v_{t+1}}{2},
\end{aligned}
\end{equation*}
where $\mathbf p_t=[x_t,y_t,z_t]^\top$,
$\mathbf v_t=[u_t,v_t,w_t]^\top$, and
$\bar\psi_t=(\psi_t+\psi_{t+1})/2$, and $R_z(\cdot)$ maps body-frame linear velocity into the inertial frame.

\subsubsection{Trajectory-conditioned policy}
\label{subsubsec:mcpilco_policy}

The tracking policy is a phase-conditioned feedback law: at each step, \(t\) selects the reference sample \(\bar{\mathbf x}^{\mathrm{ref}}_{t}\), and the policy input is built from tracking errors expressed in body-frame position coordinates and continuous yaw features. The horizontal position error is represented in the body frame:
\begin{equation*}
\mathbf e_{xy,t}^{b}
=
R_z(-\psi_t)
\left(
\mathbf p_{xy,t}^{\mathrm{ref}}-\mathbf p_{xy,t}
\right),
\qquad
e_{z,t}=z_{t}^{\mathrm{ref}}-z_t .
\end{equation*}
The yaw error is represented continuously as
\begin{equation*}
e_{\psi,t}
=
\operatorname{atan2}\!\left(
\sin(\psi_{t}^{\mathrm{ref}}-\psi_t),
\cos(\psi_{t}^{\mathrm{ref}}-\psi_t)
\right).
\end{equation*}

With $D_p^\pi$, $D_\nu^\pi$, and $D_{\nu,\mathrm{ref}}^\pi$ denoting diagonal normalization matrices, the policy feature vector is
\[
\boldsymbol{\phi}_t
=
\begin{bmatrix}
(D_p^\pi)^{-1}
\begin{bmatrix}
\mathbf e^{b}_{xy,t}\\
e_{z,t}
\end{bmatrix}
\\[1mm]
\sin e_{\psi,t}\\
\cos e_{\psi,t}\\[1mm]
(D_\nu^\pi)^{-1}
(\boldsymbol{\nu}^{\mathrm{ref}}_{t}-\boldsymbol{\nu}_t)
\\[1mm]
(D_{\nu,\mathrm{ref}}^\pi)^{-1}
\boldsymbol{\nu}^{\mathrm{ref}}_{t}
\end{bmatrix}.
\]
The first block provides local pose feedback, the second block gives a continuous yaw representation, the third block penalizes phase lag in the desired motion, and the final block acts as a feedforward control. This representation is important for non-stationary tracking: the same pose error may require different actions depending on whether the reference is accelerating, descending, turning, or moving through a coupled three-dimensional segment.

The bounded thruster command is
\begin{equation*}
\label{eq:mcpilco_policy}
\mathbf a_t
=
\pi_\theta(\mathbf x_t)
=
\mathbf a_{\max}\odot
\tanh\!\left(f_\theta(\boldsymbol\phi_t)\right),
\end{equation*}
where $f_\theta$ is a feedforward neural network and $\mathbf a_{\max}$ contains the componentwise command limits.

\subsubsection{Tracking cost}
\label{subsubsec:mcpilco_cost}

The policy parameters are optimized by minimizing the Monte Carlo finite-horizon objective
\begin{equation*}
\label{eq:mcpilco_tracking_objective}
\widehat J(\theta)
=
\sum_{t=0}^{H-1}
\frac{1}{M}
\sum_{m=1}^{M}
 c_t\!\left(\mathbf x_t^{(m)},\mathbf a_t^{(m)}\right),
\end{equation*}
where $M$ is the number of rollout particles. Define the normalized tracking errors
\begin{align*}
\tilde e_{x,t}
&=
(x_t-x^{\mathrm{ref}}_{t})/\ell_{xy},
&
\tilde e_{y,t}
&=
(y_t-y^{\mathrm{ref}}_{t})/\ell_{xy},
\\
\tilde e_{z,t}
&=
(z_t-z^{\mathrm{ref}}_{t})/\ell_z,
&
\tilde e_{\psi,t}
&=
\mathrm{wrap}(\psi_t-\psi^{\mathrm{ref}}_{t})/\ell_\psi,
\end{align*}
and
\[
\tilde{\mathbf e}_{\nu,t}
=
D_\nu^{-1}
(\boldsymbol{\nu}_t-\boldsymbol{\nu}^{\mathrm{ref}}_{t}) .
\]
To avoid excessive sensitivity to large transient errors during early model-based policy search, the tracking terms use the Huber penalty
\[
\rho_\delta(e)=
\begin{cases}
\frac{1}{2}e^2, & |e|\leq\delta,\\
\delta\left(|e|-\frac{1}{2}\delta\right), & |e|>\delta .
\end{cases}
\]
The stage cost is
\begin{equation*}
\label{eq:track_cost}
\begin{aligned}
c_t
=&\;
w_{xy}
\left[
\rho_\delta(\tilde e_{x,t})
+
\rho_\delta(\tilde e_{y,t})
\right]
+
w_z\rho_\delta(\tilde e_{z,t})
+
w_\psi\rho_\delta(\tilde e_{\psi,t})
\\
&+
w_\nu
\sum_i
\rho_\delta(\tilde e_{\nu_i,t})
+
w_u
\left\|
\mathbf a_t\oslash\mathbf a_{\max}
\right\|_2^2
\\
&+
w_{\Delta u}
\left\|
(\mathbf a_t-\mathbf a_{t-1})
\oslash
\mathbf a_{\max}
\right\|_2^2 .
\end{aligned}
\end{equation*}

The first terms penalize pose and velocity tracking errors along the reference trajectory. The final two terms regularize input magnitude and input variation, discouraging unnecessarily aggressive or oscillatory thruster commands.

The learned controller is used as the execution layer for the active certified reference. If online monitoring invalidates the remaining suffix, the controller is superseded by the hold-and-repair logic in Section~\ref{sec:method_online_replanning}. After a repaired trajectory is certified, the same policy structure tracks the updated reference.

\begin{figure*}[!t]
    \centering
    \includegraphics[width=\textwidth]{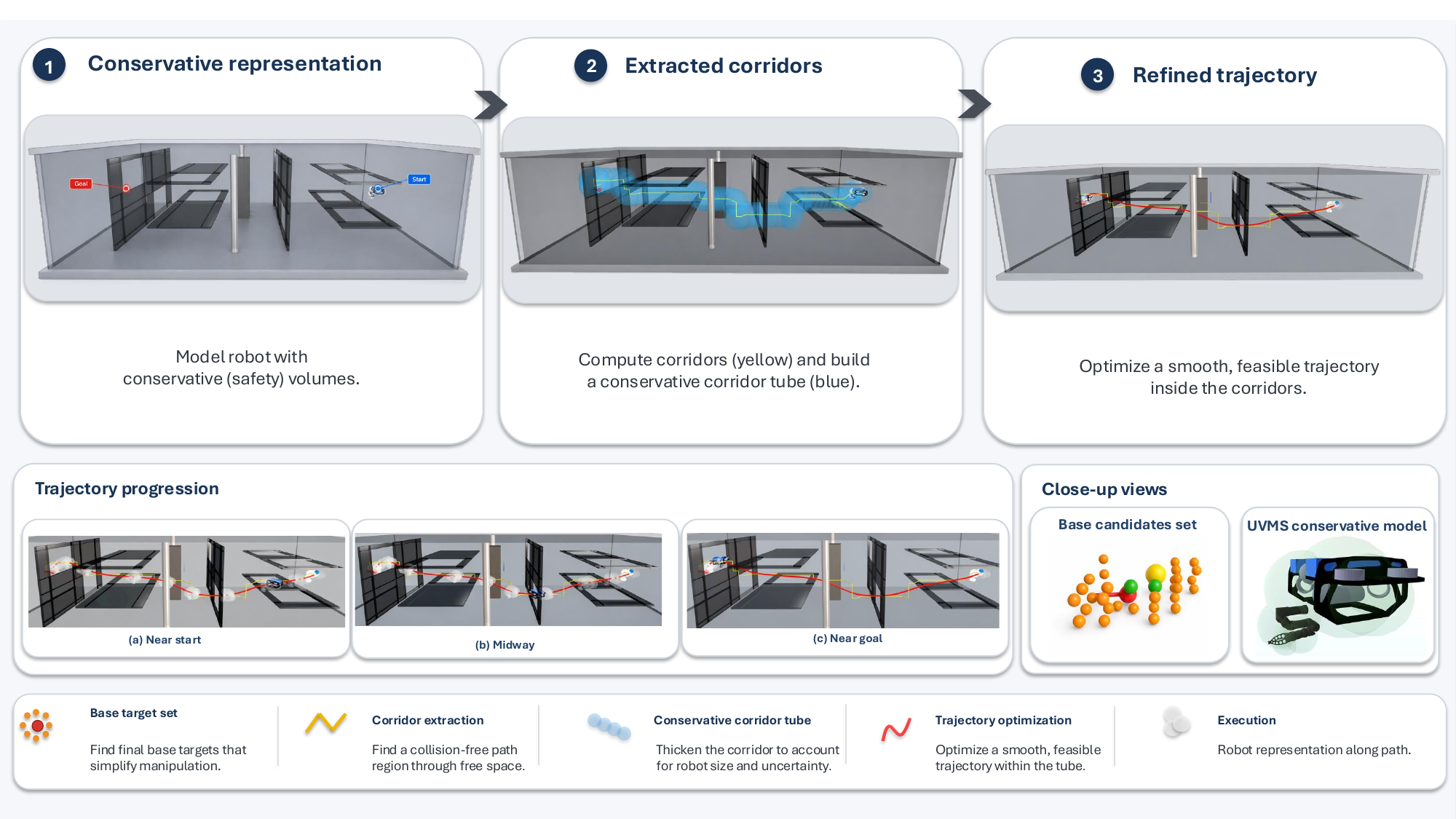}
    \caption{Planning-phase visualization: conservative model, corridor extraction, and refined trajectory.}
    \label{fig:planning_phase_env}
\end{figure*}

\section{Experimental Setup}
\label{sec:experimental_setup}

We evaluate MANTA on confined UVMS tasks in which success requires both collision-safe passage through cluttered spaces and terminal whole-body feasibility for manipulation. The experimental evaluation is divided into two parts. The first part isolates the planning problem and compares MANTA against full-state sampling-based planners over a matched set of confined-navigation queries. The second part evaluates the closed-loop execution layer by testing a learned MC-PILCO tracking policy in a ROS~2--Gazebo simulation of the BlueROV platform.

\subsection{Simulation Environment and UVMS Model}
\label{subsec:setup_sim_uvms}

The robotic platform considered in all experiments is a BlueROV-class
underwater vehicle ~\cite{BlueRobotics2024} equipped with an Alpha5 manipulator~\cite{ReachRobotics2024}. For planning, the UVMS
is modeled as a floating-base articulated system in Pinocchio~\cite{carpentier2019pinocchio}. The arm joints are \((q_1,q_2,q_3,q_4,q_5)\). During conservative reduced-space search and local
initialization, the manipulator is kept in the compact transit posture $\mathbf q_{a,\mathrm{tr}}
=
(3.05,\;1.745,\;1.745,\;2.88,\;0.0)$ rad. 

The lower-level arm optimizer acts on the three positioning joints \((q_1,q_2,q_3)\), while the remaining wrist and gripper coordinates are fixed. This choice is consistent with the planning benchmark, where the terminal task constrains the tool position rather than the full end-effector pose.

The planning maps are generated from analytic obstacle primitives and voxelized into an ESDF with \(0.08~\mathrm{m}\) resolution. Collision evaluation combines the articulated robot model with a conservative sphere approximation attached to the vehicle, arm links, and tool. For a full UVMS configuration \(\mathbf q_{f}\), the conservative clearance is $\phi(\mathbf q_f)$. A planned trajectory is considered geometrically certified only if it is collision-free under the articulated checker and satisfies
$\min_k \phi(\mathbf q_{f,k})\ge 0.01~\mathrm{m}.
$
Task success additionally requires the terminal tool-position error to be less than \(0.10~\mathrm{m}\).

Closed-loop control experiments are conducted in Gazebo simulation environment~\cite{koenig2004gazebo}. The controller operates at \(50~\mathrm{Hz}\) and commands the eight BlueROV thrusters. The learned policy is evaluated on trajectory-tracking tasks derived from tube-like references representative of confined underwater navigation.

\subsection{Benchmark Scenarios}
\label{subsec:setup_benchmarks}

The planning benchmark contains four confined environments, summarized in Table~\ref{tab:env_benchmarks}. These environments are designed to test different failure modes: laterally shifted passages, branch selection, vertical passage changes, and route-choice tradeoffs between short narrow routes and longer wider routes. Passage size is controlled by $p_\nu = 2\nu r_{\mathrm{UVMS}},
$
where \(r_{\mathrm{UVMS}}=0.325~\mathrm{m}\) and \(\nu\in\{1.2,1.4,1.8\}\). Thus the corresponding passage widths are $p_\nu\in\{0.782,\;0.912,\;1.172\}~\mathrm{m}.
$
For each environment--narrowness pair, ten start--goal queries are generated.
All planning methods and ablations are evaluated on the same \(120\) matched
queries.

\begin{table}[t!]
\centering
\caption{Benchmark environments.}
\label{tab:env_benchmarks}
\renewcommand{\arraystretch}{1.04}
\setlength{\tabcolsep}{3pt}
\scriptsize
\begin{tabularx}{\columnwidth}{@{}c c c >{\raggedright\arraybackslash}X@{}}
\toprule
\textbf{Env.} & \textbf{Branch} & \textbf{3D} & \textbf{Main challenge} \\
\midrule
E1\textsuperscript{a} & No & Limited & Laterally shifted window sequence. \\
E2\textsuperscript{b} & Yes & No & Bottlenecked upper/lower cave branches. \\
E3\textsuperscript{c} & Limited & Yes & Vertical slab openings and high/low gates. \\
E4\textsuperscript{d} & Yes & No & Short narrow route vs. longer wider route. \\
\bottomrule
\end{tabularx}
\end{table}

\begin{figure}[t]
    \centering
    \captionsetup[subfigure]{font=footnotesize}

    \begin{subfigure}[t]{0.49\columnwidth}
        \centering
        \includegraphics[width=\linewidth]{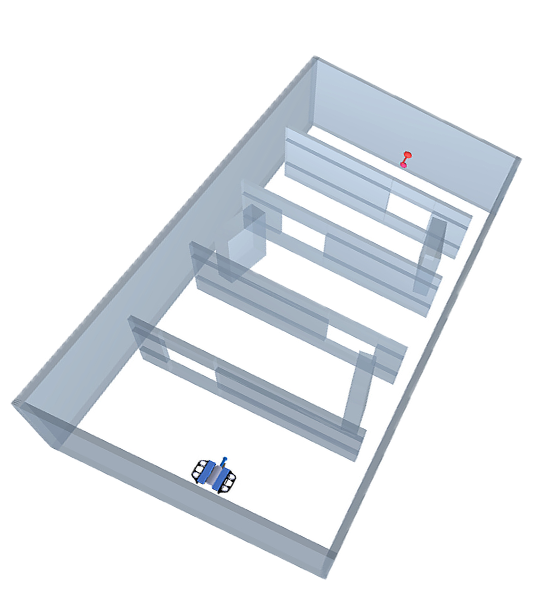}
        \caption{\texttt{S-shape}}
        \label{fig:env_view1}
    \end{subfigure}
    \hfill
    \begin{subfigure}[t]{0.49\columnwidth}
        \centering
        \includegraphics[width=\linewidth]{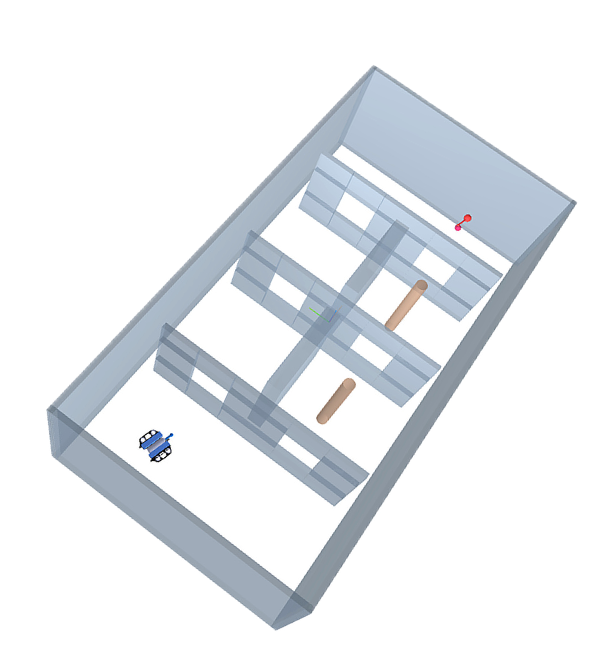}
        \caption{\texttt{Two-tunnels}}
        \label{fig:env_view2}
    \end{subfigure}

    \vspace{0.5em}

    \begin{subfigure}[t]{0.49\columnwidth}
        \centering
        \includegraphics[width=\linewidth]{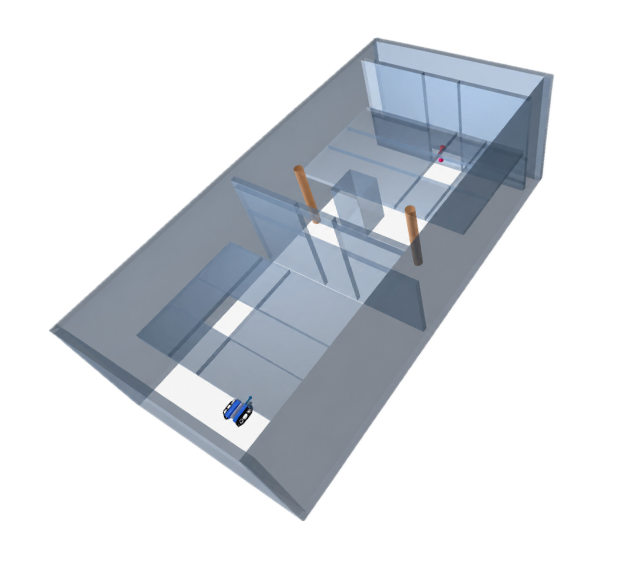}
        \caption{\texttt{Vertical-shaft}}
        \label{fig:env_view4}
    \end{subfigure}
    \hfill
    \begin{subfigure}[t]{0.49\columnwidth}
        \centering
        \includegraphics[width=\linewidth]{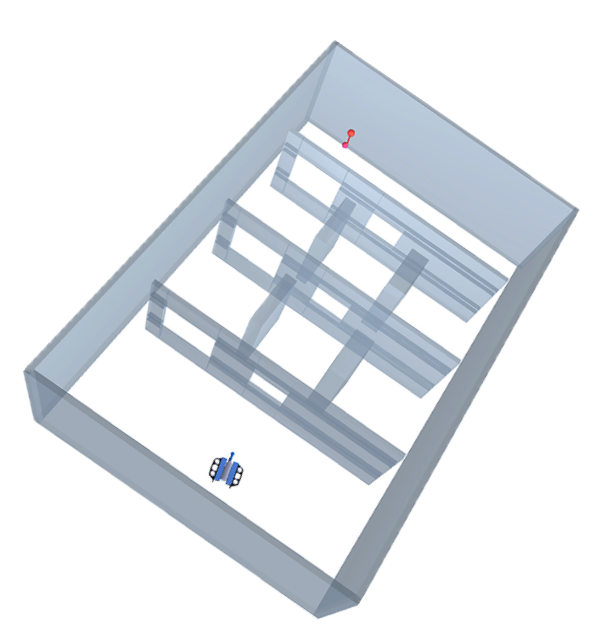}
        \caption{\texttt{Narrow Vs Wide}}
        \label{fig:env_view3}
    \end{subfigure}

    \caption{Representative confined environments.}
    \label{fig:planning_envs_onecolumn}
\end{figure}

For the control experiments, the learned policy is evaluated on one training reference and on four unseen families of tube-like trajectories: S-shaped passages, gently varying tube centerlines, vertical tube motions, and helical paths. These references excite different components of the vehicle dynamics: lateral curvature changes, depth regulation, coupled three-dimensional motion, and yaw variation.

\subsection{Baselines, Metrics, and Statistical Protocol}
\label{subsec:setup_metrics_protocol}

The planning baselines are full-state sampling-based planners implemented using the Open Motion Planning Library (OMPL)~\cite{sucan2012ompl}. We compare against RRT-Connect~\cite{Kuffner2000RRTConnect} and BiTRRT~\cite{devaurs2013enhancing}. The baseline state is
\[
\mathbf q_s=(x,y,z,\phi,\theta,\psi,q_1,q_2,q_3),
\]
where the first six variables represent the vehicle pose and the remaining variables correspond to the active arm joints. The baselines therefore search directly in a higher-dimensional planning space than MANTA's reduced global layer.

Table~\ref{tab:key_planner_params} summarizes the main MANTA and baseline planning parameters. The same query set, random seeds, collision checker, terminal-error criterion, and post-processing protocol are used for all methods.

\newcommand{\plannerparamgroup}[1]{%
    \addlinespace[2pt]
    \rowcolor{sectiongray}
    \multicolumn{2}{c}{\textbf{\sffamily\itshape #1}}\\
    \addlinespace[1pt]
}

\begin{table}[t!]
\centering
\caption{Main planning configuration.}
\label{tab:key_planner_params}
\renewcommand{\arraystretch}{1.08}
\setlength{\tabcolsep}{3pt}
\scriptsize
\begin{tabularx}{\columnwidth}{@{}>{\raggedright\arraybackslash}p{0.47\columnwidth}
>{\raggedright\arraybackslash}X@{}}
\toprule
\textbf{Parameter} & \textbf{Value} \\
\midrule

\plannerparamgroup{Global reduced search}

Reduced state \(\mathbf r\)
&
\((x,y,z,\psi)\)
\\

Lattice resolution
&
\(\Delta_{\mathrm{lat}}=0.22\,\mathrm{m}\)

\\

yaw discretization
&
\(n_\psi=12\)
\\

Corridor extraction
&
\(K_{\mathrm{pri}}=1\), \(K_{\mathrm{fb}}=2\), \(\bar\phi_{\mathrm{fb}}=0.005\,\mathrm{m}\)
\\

Safety margins
&
\(\delta_b=0.02\,\mathrm{m}\), \(\delta_{\mathrm{cert}}=0.01\,\mathrm{m}\)
\\

Goal screening
&
\(\bar e_{\mathrm{tool}}=0.16\,\mathrm{m}\), \(\bar d_{\mathrm{bt}}=1.10\,\mathrm{m}\), \(\bar e_\psi=155^\circ\)
\\

Terminal IK screening
&
\(N_{\mathrm{quick}}=8\), \(N_{\mathrm{IK}}=3\), \(I_{\mathrm{IK}}=45\)
\\

\plannerparamgroup{Upper base refinement}

Base representation
&
Cubic open B-spline \(\mathbf r(u)\)
\\

Spline size and sampling
&
\(n_c=6\)--\(24\), \(N+1=18\)--\(160\)
\\

Corridor tube and clearance shaping
&
\(r_{\mathrm{tube}}=0.45\,\mathrm{m}\), \(w_{\mathrm{clr}}=0.45\), \(\bar\phi_{\mathrm{pref}}=0.08\,\mathrm{m}\)
\\

Upper--lower coupling
&
\(\lambda_{\mathrm{low}}=0.08\)
\\

\plannerparamgroup{Lower arm refinement}

Active joints and parametrization
&
\(\mathcal J_{\mathrm{act}}=(q_1,q_2,q_3)\)
\\

Knot budget and terminal span
&
\(M_{\max}=12\), \(N_{\mathrm{term}}=5\)
\\

Critical-event thresholds
&
\(\phi_{\mathrm{crit}}=0.20\,\mathrm{m}\), \(c_{\mathrm{crit}}=0.36\,\mathrm{m}\)
\\

Main lower-level weights
&
\(w_{\mathrm{task}}=2600\), \(w_{\mathrm{cross}}=12.0\), \(w_{\mathrm{sm}}=0.80\), \(w_{\mathrm{stow}}=0.02\)
\\

\plannerparamgroup{Terminal repair}

Trigger and local segment
&
\(e_{\mathrm{repair}}=0.10\,\mathrm{m}\), \(L_{\mathrm{rep}}=6\), \(I_{\mathrm{rep}}=80\)
\\

Repair objective weights
&
\(w_{\mathrm{task}}^{\mathrm{rep}}=9000\), \(w_{\mathrm{clr}}^{\mathrm{rep}}=7000\), \(w_{\mathrm{sm}}^{\mathrm{rep}}=2.0\)
\\

Repair regularization
&
\(w_{\mathrm{arm}}^{\mathrm{rep}}=0.01\), \(w_{\psi}^{\mathrm{rep}}=0.08\),
\(b_{\mathrm{clr}}^{\mathrm{rep}}=0.015\,\mathrm{m}\),
\(\Delta\psi_{\mathrm{rep}}=18^\circ\)
\\

\plannerparamgroup{Sampling baselines}

State
&
\(\mathbf q_s=(x,y,z,\phi,\theta,\psi,q_1,q_2,q_3)\)
\\

Planners
&
RRT-Connect and BiTRRT
\\

Sampling limits
&
Iterations \(20000/25000\), goal bias \(0.08\)
\\

Post-processing and checking
&
Check/interp. step \(0.035\,\mathrm{m}\), \(w_a=0.25\), \(T_0=1.0\)
\\

\bottomrule
\end{tabularx}
\end{table}

The primary planning metric is task success, defined as simultaneous geometric certification and terminal tool-position error below \(0.10~\mathrm{m}\). We also report geometric success, wall-clock planning time, minimum conservative clearance, terminal tool error, base path length, and total arm motion. Minimum clearance is particularly important in confined environments because it quantifies the margin available to absorb modeling errors, tracking deviations, and map discretization effects. Smoothness is evaluated using base curvature energy, arm total variation, maximum arm step, and second-difference energies. These are planning-level regularity indicators and are not treated as closed-loop tracking guarantees.

For control, the main metrics are three-dimensional position
\(\mathrm{RMSE}_p\), horizontal and vertical components
\(\mathrm{RMSE}_{xy}\) and \(\mathrm{RMSE}_z\), yaw
\(\mathrm{RMSE}_\psi\), mean position error, maximum position error, RMS thrust magnitude, and thrust saturation percentage.

The MC-PILCO policy is trained from an initial dataset collected using a PD controller. The proportional position gains are $K_p=[30,\;30,\;45],$
and the derivative gains are $K_d=[20,\;20,\;35,\;3,\;3,\;2],$ with attitude gain \(6.5\). All rollouts are sampled at \(50~\mathrm{Hz}\) over a \(30~\mathrm{s}\) horizon.
The learned policy outputs the eight BlueROV thruster commands and uses the componentwise bound $\mathbf a_{\max} = [10,\;10,\;10,\;10,\;10,\;10,\;10,\;10]^\top .$

The policy is a feedforward neural network with two hidden layers of \(32\) and \(64\) units and ReLU activations. Policy optimization uses \(1200\) Monte Carlo particles, \(2500\) gradient steps, and learning rate \(3\times10^{-4}\). The GP model predicts the velocity increments \((\Delta u,\Delta v,\Delta w,\Delta r)\) using RBF kernels. GP hyperparameters are optimized for \(1500\) steps using mini-batches of \(500\) samples, and a subset-of-regressors approximation with a maximum of \(1000\) inducing points is used. The learning process is run for six closed-loop trials.

\begin{table}[t!]
\centering
\caption{Main MC-PILCO tracking configuration.}
\label{tab:mcpilco_training_setup}
\footnotesize
\setlength{\tabcolsep}{4.5pt}
\renewcommand{\arraystretch}{1.12}
\begin{tabular}{lcc}
\toprule
\textbf{Quantity} & \textbf{Symbol} & \textbf{Value} \\
\midrule
Sampling frequency & \(f_s\) & \(50\,\mathrm{Hz}\) \\
Prediction horizon & \(T\) & \(30\,\mathrm{s}\) \\
Number of learning trials & \(N_{\mathrm{trial}}\) & \(6\) \\
Policy hidden layers & $h_1, h_2$ & \(32,\;64\) \\
Number of rollout particles & \(M\) & \(1200\) \\
Policy optimization steps & \(N_{\theta}\) & \(2500\) \\
Policy learning rate & \(\alpha_{\theta}\) & \(3\times10^{-4}\) \\
GP optimization steps & \(N_{\mathrm{GP}}\) & \(1500\) \\
Mini-batch size & \(B\) & \(500\) \\
SoR regressors & \(M_{\mathrm{SoR}}\) & \(1000\) \\
Particle initialization pieces & \(N_{\mathrm{seg}}\) & \(10\) \\
\bottomrule
\end{tabular}
\end{table}

Success rates are reported as percentages over the matched query set. Continuous planning metrics are reported as median \([Q_1,Q_3]\), where \(Q_1\) and \(Q_3\) denote the 25th and 75th percentiles. Control generalization metrics are reported as mean \(\pm\) standard deviation over ten trajectories per family.

\section{Results}
\label{sec:results}

\subsection{Planning Experiments}
\label{subsec:planning_results}

The main comparison over the \(120\) matched planning queries is reported in Table~\ref{tab:overall_method_summary}. MANTA achieves \(92.5\%\) task success, whereas RRT-Connect and BiTRRT achieve \(70.8\%\) and \(70.0\%\), respectively. The sampling baselines are geometrically successful in all trials, but they frequently violate the terminal tool-position criterion. This shows that collision-free connectivity alone is insufficient for confined UVMS manipulation: the planner must also select a terminal operating state that supports the downstream arm task.

\begin{table*}[t!]
\centering
\caption{Overall planning comparison over 120 matched queries. Success rates are
percentages. Continuous metrics are reported as median \([Q_1,Q_3]\).}
\label{tab:overall_method_summary}
\renewcommand{\arraystretch}{1.05}
\setlength{\tabcolsep}{4pt}
\scriptsize
\begin{tabularx}{\textwidth}{@{}l c c c c c c c@{}}
\toprule
\textbf{Method} & \textbf{Task [\%]} & \textbf{Geom. [\%]} & \textbf{Time [s]} &
\textbf{Min clr. [m]} & \textbf{Term. err. [m]} & \textbf{Base [m]} & \textbf{Arm [rad]} \\
\midrule
MANTA (ours)
& \textbf{92.5}
& 99.2
& \(8.55\,[7.35,\,11.15]\)
& \(\mathbf{0.100\,[0.065,\,0.133]}\)
& \(0.013\,[0.001,\,0.068]\)
& \(\mathbf{10.51\,[9.80,\,12.58]}\)
& \(\mathbf{1.11\,[0.70,\,2.96]}\)
\\
RRT-Connect
& 70.8
& {100.0}
& \(1.37\,[0.91,\,2.62]\)
& \(0.015\,[0.012,\,0.026]\)
& \({0.000\,[0.000,\,0.151]}\)
& \(10.89\,[9.78,\,11.89]\)
& \(3.84\,[1.98,\,5.51]\)
\\
BiTRRT
& 70.0
& {100.0}
& \({1.27\,[0.87,\,2.95]}\)
& \(0.015\,[0.012,\,0.030]\)
& \({0.000\,[0.000,\,0.187]}\)
& \(10.64\,[9.77,\,11.95]\)
& \(3.30\,[1.81,\,5.04]\)
\\
\bottomrule
\end{tabularx}
\end{table*}

Beyond binary success, the most important difference is the clearance margin. MANTA obtains a median minimum clearance of \(0.100~\mathrm{m}\), compared with \(0.015~\mathrm{m}\) for both RRT-Connect and BiTRRT. This larger clearance is critical in confined environments, where small tracking errors, unmodeled hydrodynamic disturbances, discretization artifacts, or perception uncertainty can turn near-contact plans into unsafe executions. The sampling baselines are faster in median wall time, but they tend to return paths that pass close to obstacles and require larger manipulator reconfiguration. In contrast, MANTA uses the reduced global route, clearance-aware refinement, and terminal repair to produce plans with higher safety margins and lower arm motion.

\begin{figure*}[t!]
\centering
\includegraphics[width=0.245\textwidth]{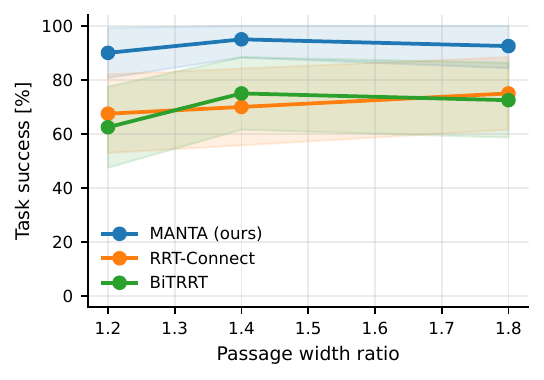}
\includegraphics[width=0.245\textwidth]{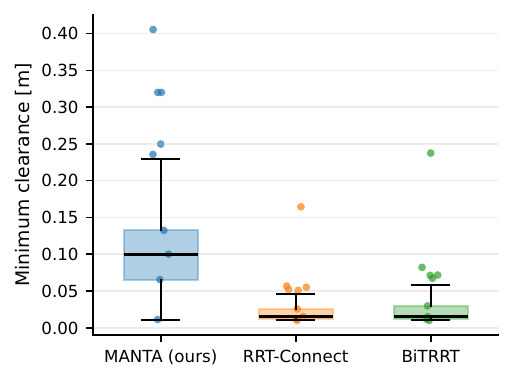}
\includegraphics[width=0.245\textwidth]{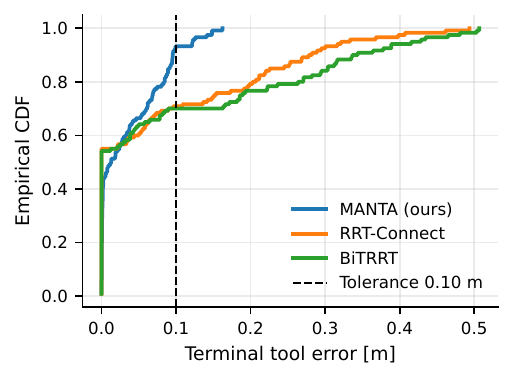}
\includegraphics[width=0.245\textwidth]{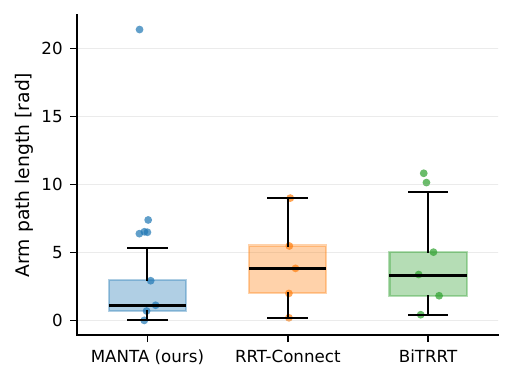}
\caption{Planning comparison with full-state sampling-based baselines. From
left to right: task success versus narrowness, minimum conservative clearance,
terminal tool-error distribution, and total arm motion.}
\label{fig:baseline_core_results}
\end{figure*}

The four panels in Fig.~\ref{fig:baseline_core_results} provide complementary views of the same comparison. The success plot shows how reliability varies with passage narrowness, the clearance boxplot emphasizes safety margin, the terminal-error CDF reveals the dominant failure mode of the sampling baselines, and the arm-motion distribution shows the cost of searching directly in the full state space.

At the environment level, MANTA obtains \(100.0\%\) task success in E1, \(96.7\%\) in E3, and \(86.7\%\) in E2 and E4. The lower success in the branching maps is expected because these environments impose stronger coupling between route-family selection and terminal manipulation feasibility. The dominant baseline failure mode is not geometric collision, but residual terminal tool error, especially in bottlenecked and route-choice maps.

The planning-level smoothness metrics support the same interpretation. On successful trajectories, MANTA achieves lower median arm total variation than RRT-Connect and BiTRRT: \(1.068~\mathrm{rad}\), compared with
\(3.327~\mathrm{rad}\) and \(2.473~\mathrm{rad}\), respectively. It also
substantially reduces the median base curvature energy, from
\(0.3818~\mathrm{rad}^2/\mathrm{m}\) for RRT-Connect and
\(0.3276~\mathrm{rad}^2/\mathrm{m}\) for BiTRRT to
\(0.0368~\mathrm{rad}^2/\mathrm{m}\). These results indicate that MANTA produces smoother base motions and avoids unnecessary arm reconfiguration on
successful plans.

The maximum arm step is larger for MANTA because the arm is often kept compact during transit and deployed near the terminal task. Therefore, these metrics should be interpreted as evidence of improved planning-level regularity, not as a substitute for closed-loop tracking evaluation.

Table~\ref{tab:overall_ablation_summary} reports the ablation study over the same 120 matched queries. We want to show how individual design choices affect task feasibility, clearance, terminal error, and arm motion.

\begin{table*}[t!]
\centering
\caption{Ablation summary over 120 matched queries. Success rates are percentages;
other metrics are median \([Q_1,Q_3]\).}
\label{tab:overall_ablation_summary}
\renewcommand{\arraystretch}{1.05}
\setlength{\tabcolsep}{5pt}
\scriptsize
\begin{tabular}{@{}l c c c c c@{}}
\toprule
\textbf{Variant} & \textbf{Task [\%]} & \textbf{Geom. [\%]} &
\textbf{Min clr. [m]} & \textbf{Term. err. [m]} & \textbf{Arm [rad]} \\
\midrule
MANTA (ours)
& 92.5
& \textbf{99.2}
& \(\mathbf{0.100\,[0.065,\,0.133]}\)
& \(\mathbf{0.013\,[0.001,\,0.068]}\)
& \(1.11\,[0.70,\,2.96]\) \\
w/o terminal repair
& 40.8
& 86.7
& \(0.083\,[0.034,\,0.136]\)
& \(0.107\,[0.079,\,0.135]\)
& \(0.000\,[0.000,\,0.591]\) \\
w/o upper clearance
& 86.7
& 88.3
& \(0.076\,[0.032,\,0.133]\)
& \(0.045\,[0.000,\,0.082]\)
& \(0.591\,[0.000,\,1.362]\) \\
w/o cross-section
& 88.3
& 90.0
& \(0.080\,[0.033,\,0.133]\)
& \(0.034\,[0.000,\,0.075]\)
& \(0.578\,[0.000,\,1.124]\) \\
w/o arm surrogate
& 88.3
& 90.0
& \(0.082\,[0.034,\,0.136]\)
& \(0.044\,[0.000,\,0.082]\)
& \(0.654\,[0.000,\,1.570]\) \\
w/o stow/smoothness
& 95.8
& 98.3
& \(0.134\,[0.094,\,0.160]\)
& \(0.037\,[0.000,\,0.076]\)
& \(0.803\,[0.000,\,1.585]\) \\
\bottomrule
\end{tabular}
\end{table*}

The ablation results indicate that terminal repair is the dominant component for task-level reliability. Without it, geometric success remains relatively high, but task success drops to \(40.8\%\), and the median terminal error increases to \(0.107~\mathrm{m}\), above the task threshold. This confirms that many failures arise after a collision-free route has already been found: the remaining challenge is to adjust the terminal arm and yaw configuration so that the tool reaches the target accurately.

Upper-level clearance shaping has a smaller but meaningful effect. Removing it reduces the median clearance and slightly lowers task success, which is consistent with its role as a safety-margin term rather than a direct terminal accuracy mechanism. The cross-section and arm-surrogate ablations have weaker aggregate effects under the final tuning. This does not imply that these terms are unnecessary; rather, in this benchmark their contribution is partly masked by the conservative lattice posture, terminal repair, and exact certification stage. Finally, the stow/smoothness ablation should be interpreted carefully. Although removing these regularizers increases raw success, it weakens the intended behavior of keeping the arm compact during transit and reducing unnecessary arm motion. We therefore treat these terms as trajectory-behavior regularizers. 

\begin{figure*}[t!]
\centering
\includegraphics[width=0.5\textwidth]{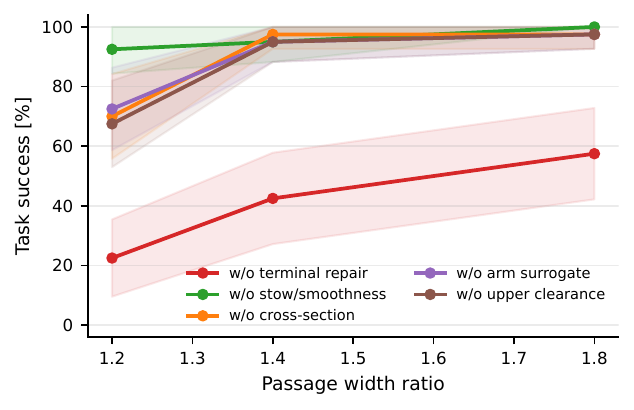}
\includegraphics[width=0.37\textwidth]{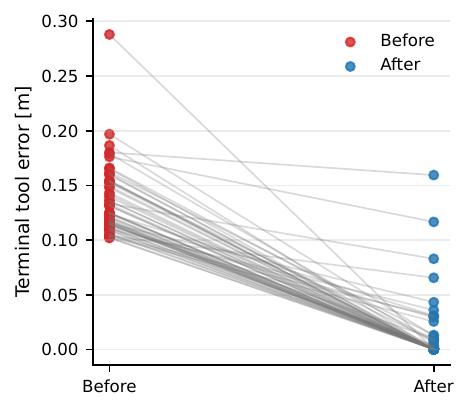}
\caption{Ablation results. Left: task success versus passage narrowness for
the ablated variants. Right: effect of terminal repair on terminal task
satisfaction.}
\label{fig:ablation_core_results}
\end{figure*}

The ablation plots in Fig.~\ref{fig:ablation_core_results} isolate two effects that are not fully visible from the aggregate table: how success degrades with passage narrowness, and how terminal repair changes the terminal task-error distribution.

Overall, the planning experiments support three conclusions. First, MANTA improves task-level reliability compared with full-state sampling-based planning. Second, the advantage is not only binary success: MANTA produces larger clearance margins and less unnecessary arm motion, both of which are important for confined underwater operation. Third, terminal repair is the most important component for transforming a geometrically feasible route into a manipulation-feasible UVMS plan, while clearance shaping and arm regularization mainly affect safety margin and trajectory behavior.

\subsection{Control Experiments}
\label{subsec:control_results}

The control experiments evaluate whether the MC-PILCO policy can track planned-like references more accurately than the PD controller used to collect the initial data. The evaluation has two stages. First, we compare the PD rollout, the first learned policy, and the final policy after six learning trials on the training reference. Second, we apply the final policy, without additional learning, to forty unseen references grouped into four tube-like trajectory families.

\begin{figure}[t!]
\centering
\includegraphics[width=\columnwidth]{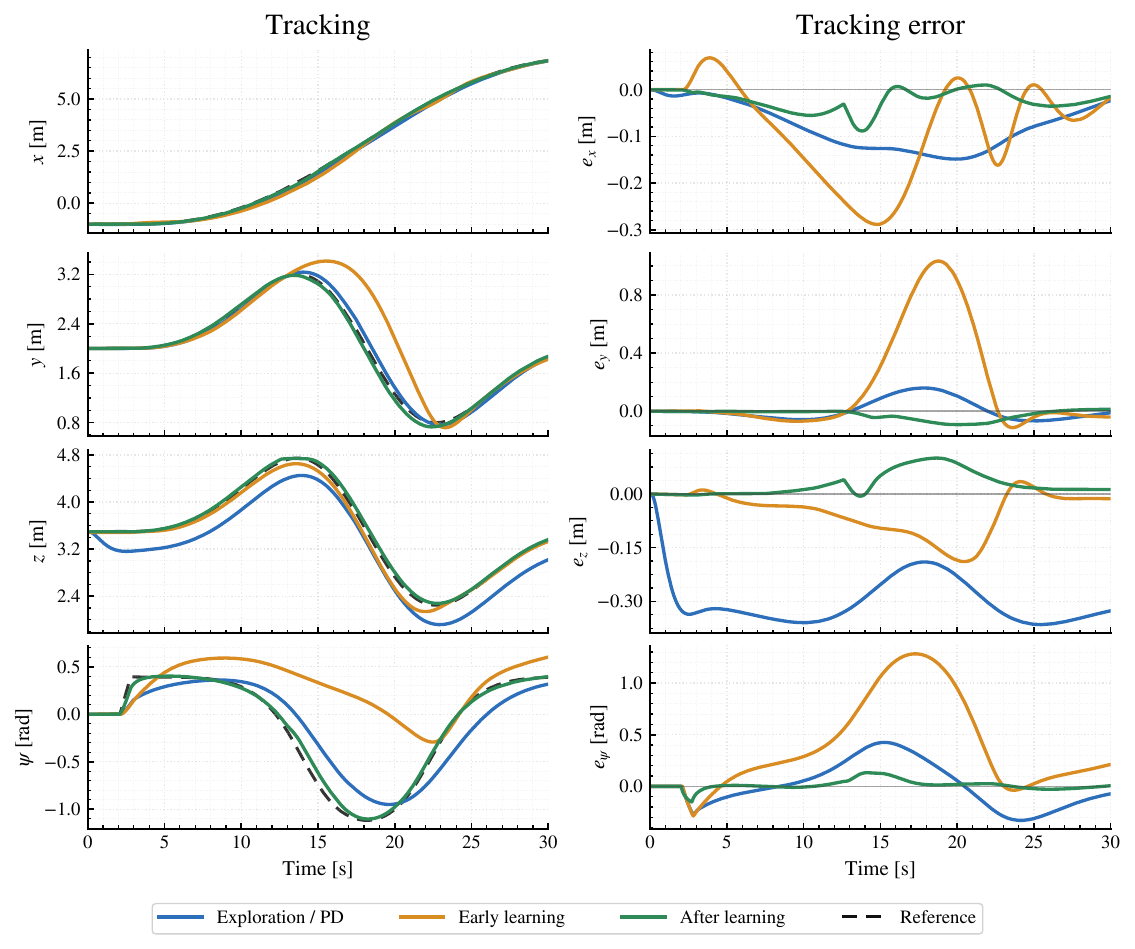}
\caption{Tracking response on the training reference for the PD exploration
controller, the first learned MC-PILCO policy, and the selected policy after
six learning trials.}
\label{fig:training_tracking_comparison}
\end{figure}

On the training reference, the learned controller substantially reduces the dominant tracking errors. The comparison in Fig.~\ref{fig:training_tracking_comparison} shows that the PD controller follows the overall trend of the reference but exhibits a persistent position bias, especially in depth. The first learned policy improves the final position error but produces a larger trajectory-wide deviation, which is consistent with policy optimization through an initially imperfect GP model. After six closed-loop trials, the learned policy tracks the reference much more closely over the full horizon.

The quantitative results in Table~\ref{tab:training_reference_results} confirm the improvement. MC-PILCO reduces the position RMSE from \(0.329~\mathrm{m}\) to \(0.065~\mathrm{m}\), corresponding to an \(80.2\%\) reduction relative to the PD baseline. The mean position error decreases by \(83.7\%\), the maximum position error by \(65.1\%\), and the yaw RMSE by \(77.9\%\). The early learned policy is useful diagnostically: despite having a smaller final error than PD, its larger RMSE and peak error show that terminal accuracy alone is not sufficient for confined-passage execution. The vehicle must remain close to the reference throughout the trajectory.

\begin{table}[t!]
\centering
\caption{Tracking performance on the training reference.}
\label{tab:training_reference_results}
\footnotesize
\setlength{\tabcolsep}{4.0pt}
\renewcommand{\arraystretch}{1.12}
\begin{tabular}{lcccc}
\toprule
\textbf{Controller} &
\textbf{RMSE$_p$} &
\textbf{Mean $e_p$} &
\textbf{Max. $e_p$} &
\textbf{RMSE$_\psi$} \\
& \textbf{[m]} & \textbf{[m]} & \textbf{[m]} & \textbf{[rad]} \\
\midrule
PD baseline & 0.329 & 0.324 & 0.378 & 0.209 \\
Early MC-PILCO, T1 & 0.404 & 0.260 & 1.045 & 0.582 \\
MC-PILCO, T6 & \textbf{0.065} & \textbf{0.053} & \textbf{0.132} & \textbf{0.046} \\
\midrule
Improvement over PD & \textbf{80.2\%} & \textbf{83.7\%} & \textbf{65.1\%} & \textbf{77.9\%} \\
\bottomrule
\end{tabular}
\end{table}

\begin{figure}[t!]
\centering
\includegraphics[width=\columnwidth]{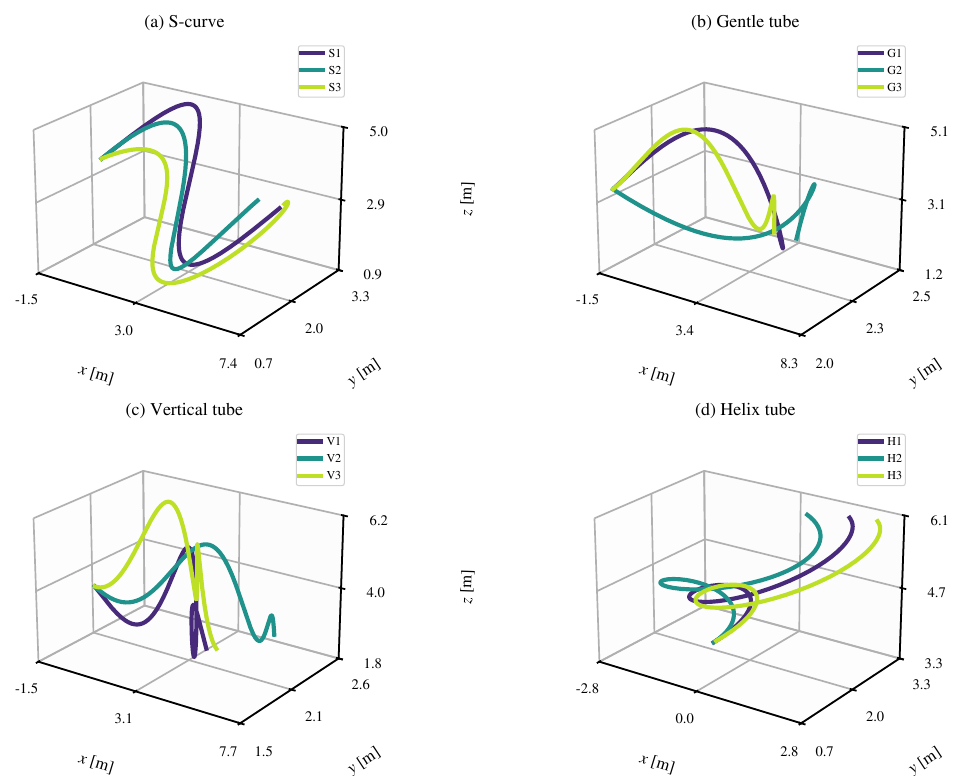}
\caption{Representative unseen trajectory families used to evaluate policy
generalization.}
\label{fig:trajectory_family_samples}
\end{figure}

Generalization is evaluated on \(40\) unseen trajectories, with ten references per family. The representative references in Fig.~\ref{fig:trajectory_family_samples} include S-curves, gentle tube centerlines, vertical tube motions, and helices. These trajectories are geometrically related to confined UVMS operation but are not used for policy optimization.

Table~\ref{tab:generalization_tracking_results} reports family-level generalization performance. Across all unseen references, MC-PILCO reduces the aggregate position RMSE from \(0.334~\mathrm{m}\) to \(0.066~\mathrm{m}\), corresponding to an \(80.1\%\) reduction. The strongest and most consistent improvement occurs in the vertical channel: \(\mathrm{RMSE}_z\) decreases from \(0.317~\mathrm{m}\) to \(0.032~\mathrm{m}\). This result is especially relevant for confined underwater environments, where depth errors directly consume clearance from the upper or lower passage boundaries.

\begin{table*}[t!]
\centering
\caption{Generalization performance on trajectory families. Values are mean
\(\pm\) standard deviation over ten trajectories per family.}
\label{tab:generalization_tracking_results}
\footnotesize
\setlength{\tabcolsep}{5.5pt}
\renewcommand{\arraystretch}{1.15}
\begin{tabular}{llccccc}
\toprule
\multicolumn{1}{c}{\multirow{2}{*}{\textbf{Family}}}
& \multicolumn{1}{c}{\multirow{2}{*}{\textbf{Controller}}}
& \multicolumn{1}{c}{\textbf{RMSE$_p$}}
& \multicolumn{1}{c}{\textbf{RMSE$_{xy}$}}
& \multicolumn{1}{c}{\textbf{RMSE$_z$}}
& \multicolumn{1}{c}{\textbf{RMSE$_\psi$}}
& \multicolumn{1}{c}{\textbf{Max. $e_p$}} \\
& & \textbf{[m]} & \textbf{[m]} & \textbf{[m]} & \textbf{[rad]} & \textbf{[m]} \\
\midrule
\multirow{2}{*}{S-curve}
& PD baseline
& \(0.336\pm0.010\) & \(0.120\pm0.024\) & \(0.313\pm0.003\) & \(0.231\pm0.057\) & \(0.471\pm0.076\) \\
& MC-PILCO
& \(\mathbf{0.068\pm0.013}\) & \(\mathbf{0.057\pm0.013}\) & \(\mathbf{0.038\pm0.007}\) & \(\mathbf{0.068\pm0.027}\) & \(\mathbf{0.134\pm0.033}\) \\
\midrule
\multirow{2}{*}{Gentle tube}
& PD baseline
& \(0.327\pm0.002\) & \(0.096\pm0.007\) & \(0.313\pm0.003\) & \(0.045\pm0.030\) & \(0.401\pm0.034\) \\
& MC-PILCO
& \(\mathbf{0.061\pm0.005}\) & \(\mathbf{0.049\pm0.006}\) & \(\mathbf{0.035\pm0.006}\) & \(\mathbf{0.038\pm0.012}\) & \(\mathbf{0.120\pm0.017}\) \\
\midrule
\multirow{2}{*}{Vertical tube}
& PD baseline
& \(0.334\pm0.006\) & \(0.097\pm0.010\) & \(0.319\pm0.008\) & \(0.072\pm0.040\) & \(0.508\pm0.064\) \\
& MC-PILCO
& \(\mathbf{0.083\pm0.018}\) & \(\mathbf{0.072\pm0.020}\) & \(\mathbf{0.039\pm0.004}\) & \(\mathbf{0.054\pm0.019}\) & \(\mathbf{0.192\pm0.048}\) \\
\midrule
\multirow{2}{*}{Helix tube}
& PD baseline
& \(0.339\pm0.009\) & \(0.094\pm0.028\) & \(0.325\pm0.002\) & \(0.449\pm0.069\) & \(0.394\pm0.033\) \\
& MC-PILCO
& \(\mathbf{0.054\pm0.022}\) & \(\mathbf{0.052\pm0.022}\) & \(\mathbf{0.014\pm0.003}\) & \(\mathbf{0.099\pm0.019}\) & \(\mathbf{0.121\pm0.053}\) \\
\midrule
\multirow{2}{*}{All families}
& PD baseline
& \(0.334\pm0.009\) & \(0.102\pm0.022\) & \(0.317\pm0.007\) & \(0.199\pm0.170\) & \(0.443\pm0.072\) \\
& MC-PILCO
& \(\mathbf{0.066\pm0.018}\) & \(\mathbf{0.057\pm0.018}\) & \(\mathbf{0.032\pm0.012}\) & \(\mathbf{0.065\pm0.030}\) & \(\mathbf{0.141\pm0.049}\) \\
\bottomrule
\end{tabular}
\end{table*}

\begin{figure}[t!]
\centering
\includegraphics[width=\columnwidth]{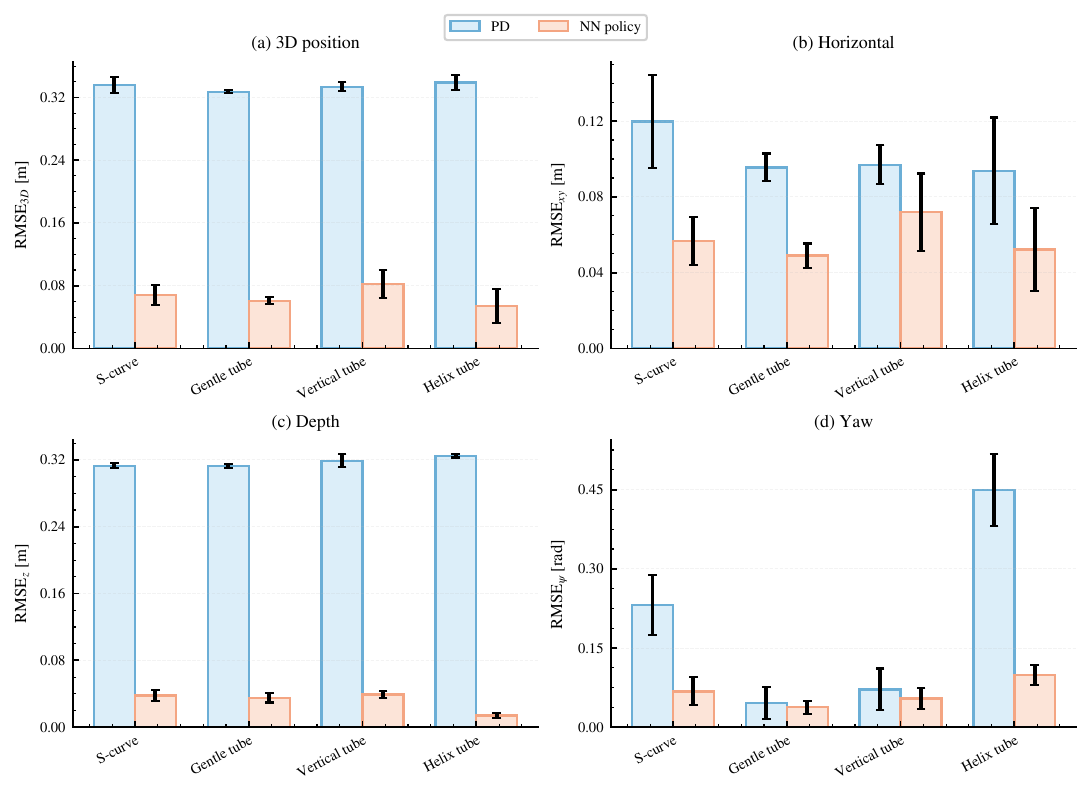}
\caption{Family-level tracking summary over unseen references.}
\label{fig:family_tracking_summary}
\end{figure}

The family-level summary in Fig.~\ref{fig:family_tracking_summary} highlights that the learned controller does not improve all families uniformly. The helix family is the most demanding case because it couples horizontal translation, vertical motion, and yaw variation; even there, MC-PILCO reduces \(\mathrm{RMSE}_p\) from \(0.339~\mathrm{m}\) to \(0.054~\mathrm{m}\) and yaw RMSE from \(0.449~\mathrm{rad}\) to \(0.099~\mathrm{rad}\). For the vertical-tube family, the primary benefit is depth regulation, whereas for the gentle-tube family the yaw improvement is smaller because the PD baseline is already accurate in heading.

\begin{figure}[t!]
\centering
\includegraphics[width=\columnwidth]{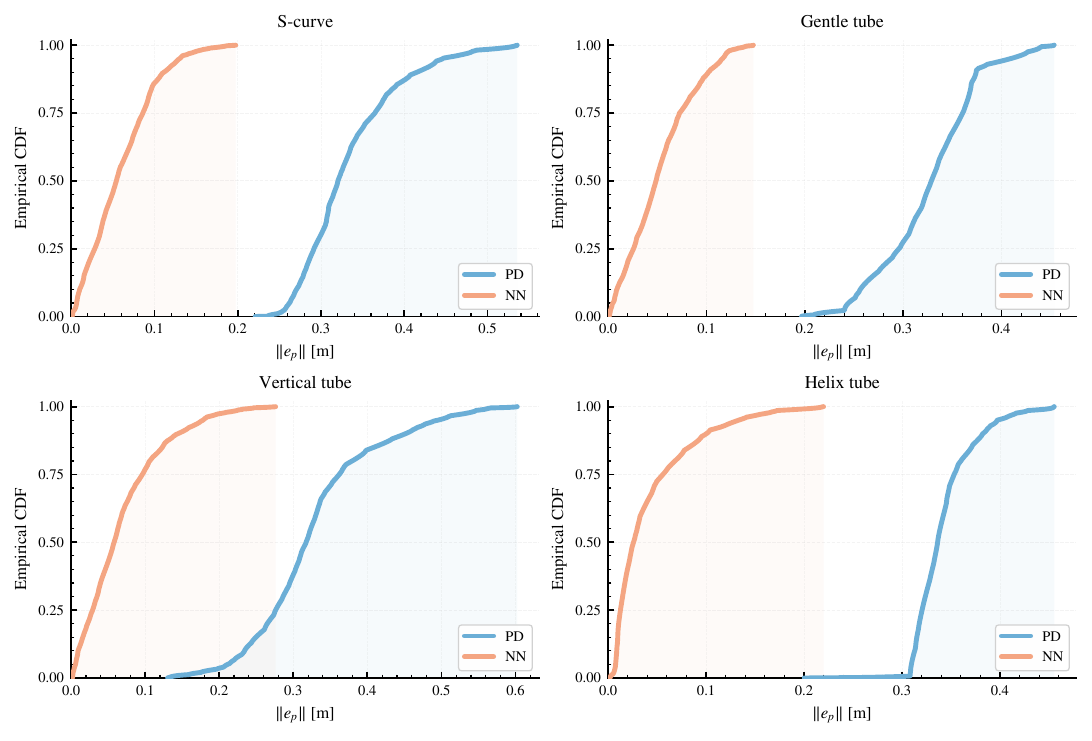}
\caption{Empirical cumulative distribution of position errors for the unseen
trajectory families.}
\label{fig:position_error_cdf}
\end{figure}

\begin{table}[!t]
\centering
\caption{Paired tracking comparison on unseen trajectories. Each entry reports
the number of trajectories where MC-PILCO outperforms the PD baseline.}
\label{tab:generalization_tracking_wins}
\footnotesize
\setlength{\tabcolsep}{5.0pt}
\renewcommand{\arraystretch}{1.15}
\begin{tabular}{lccc}
\toprule
\textbf{Family} &
\textbf{RMSE$_p$} &
\textbf{RMSE$_\psi$} &
\textbf{Max. $e_p$} \\
\midrule
S-curve       & \(\mathbf{10/10}\) & \(\mathbf{10/10}\) & \(\mathbf{10/10}\) \\
Gentle tube   & \(\mathbf{10/10}\) & \(5/10\)            & \(\mathbf{10/10}\) \\
Vertical tube & \(\mathbf{10/10}\) & \(5/10\)            & \(\mathbf{10/10}\) \\
Helix tube    & \(\mathbf{10/10}\) & \(\mathbf{10/10}\) & \(\mathbf{10/10}\) \\
\midrule
All families  & \(\mathbf{40/40}\) & \(\mathbf{30/40}\) & \(\mathbf{40/40}\) \\
\bottomrule
\end{tabular}
\end{table}


\begin{table*}[t!]
\centering
\caption{Control-signal characteristics on trajectory families. RMS entries are reported as
PD/MC-PILCO.}
\label{tab:control_signal_characteristics}
\footnotesize
\setlength{\tabcolsep}{5.0pt}
\renewcommand{\arraystretch}{1.15}
\begin{tabular}{lcccc}
\toprule
\multirow{2}{*}{\textbf{Family}} &
\multicolumn{2}{c}{\textbf{Command magnitude}} &
\textbf{Effort ratio} &
\textbf{Relative variation} \\
\cmidrule(lr){2-3}
& RMS \(u\) & Sat. [\%] & MC/PD & RMS \(\Delta u\)/RMS \(u\) \\
\midrule
S-curve
& \(1.893/2.032\)
& \(0.008/0.000\)
& \(1.074\)
& \(0.021/0.059\) \\
Gentle tube
& \(1.583/1.483\)
& \(0.000/0.000\)
& \(0.937\)
& \(0.017/0.082\) \\
Vertical tube
& \(1.776/1.793\)
& \(0.003/0.000\)
& \(1.010\)
& \(0.052/0.061\) \\
Helix tube
& \(\mathbf{18.736/2.156}\)
& \(\mathbf{0.411/0.021}\)
& \(\mathbf{0.115}\)
& \(\mathbf{1.115/0.071}\) \\
\midrule
Non-helix avg.
& \(1.750/1.770\)
& \(0.003/0.000\)
& \(1.011\)
& \(0.030/0.066\) \\
\midrule
All families
& \(\mathbf{5.997/1.866}\)
& \(\mathbf{0.105/0.005}\)
& \(\mathbf{0.311}\)
& \(\mathbf{0.878/0.068}\) \\
\bottomrule
\end{tabular}
\end{table*}
A distributional view is given in Fig.~\ref{fig:position_error_cdf}. The empirical CDFs shift toward smaller position errors for all trajectory families, indicating that the improvement is not caused by a few favorable trajectories. This is important for confined execution, where rare large deviations can be more safety-critical than the average error.

The paired comparison in Table~\ref{tab:generalization_tracking_wins} confirms the consistency of the improvement. MC-PILCO obtains lower position RMSE and lower maximum position error on all \(40\) unseen trajectories. Yaw RMSE is reduced in \(30\) out of \(40\) cases, with the non-improving cases mainly occurring in families where the PD yaw error is already small.

Finally, Table~\ref{tab:control_signal_characteristics} separates tracking accuracy from input behavior. RMS \(u\) measures average thrust effort, while RMS \(\Delta u\)/RMS \(u\) measures relative sample-to-sample command variation. For the S-curve, gentle-tube, and vertical-tube families, the MC-PILCO effort remains close to the PD baseline; the non-helical average effort ratio is \(1.011\). Thus, the learned controller does not obtain its tracking advantage by simply increasing thrust magnitude. The helix case is qualitatively different: the PD controller produces large command excursions, whereas MC-PILCO reduces both RMS effort and relative command variation while improving tracking.

The control experiments support two conclusions. First, MC-PILCO substantially improves trajectory tracking with only six closed-loop learning trials. The improvement is not limited to the terminal state; the learned policy reduces persistent position bias and improves yaw tracking along the trajectory. Second, the selected policy transfers to geometrically related unseen references. The most consistent improvement appears in depth regulation, which is directly relevant to maintaining clearance in confined underwater passages.

These results should nevertheless be interpreted as empirical tracking evidence rather than a formal safety guarantee. The learned policy improves centerline tracking and reduces large deviations, both of which support execution of clearance-aware plans. However, obstacle avoidance, actuator constraints beyond saturation, and formal safety invariance are not encoded directly in the learned policy. A deployment-oriented implementation should therefore combine the learned tracker with a supervisory safety mechanism, such as a control barrier function, model-predictive safety filter, or corridor-aware reference governor.

\section{Conclusion}
\label{sec:conclusion}

This paper presented MANTA, a hierarchical planning-and-control framework for UVMS intervention in confined underwater environments. The results show that collision-free access alone is not sufficient for manipulation: the vehicle must also preserve clearance, avoid unnecessary arm motion, and terminate in a whole-body state from which the task is feasible. MANTA addresses these requirements by combining reduced-space passage search, corridor-conditioned base refinement, arm trajectory optimization, terminal repair, and learned reach-and-hold execution.

Across 120 matched planning queries, MANTA achieved higher task success than full-state sampling-based baselines while producing larger clearance margins and lower arm motion. The ablation study showed that terminal repair is the dominant component for converting geometric feasibility into task feasibility, whereas clearance shaping and arm regularization mainly improve safety margins and trajectory quality. Closed-loop experiments further showed that an MC-PILCO policy can track planned-like references more accurately than the PD controller used for data collection. After only six learning trials, the policy reduced position and yaw errors on both the training reference and generalization trajectories, with particularly consistent improvement in depth regulation.

The present study evaluates geometric certification using conservative ESDF-based maps and exact articulated collision checking. This certification is not inherently restricted to simulation, provided that a sufficiently reliable ESDF or equivalent geometric map is available on the real system. However, the learned controller is evaluated empirically and does not by itself provide formal safety guarantees under perception errors, actuator constraints beyond saturation, or hydrodynamic uncertainty. Future work will therefore integrate MANTA with supervisory safety mechanisms such as control barrier functions, or model-predictive safety filters. Further directions include hardware validation on real UVMS platforms in confined underwater environments.

\section*{Acknowledgment}


\bibliographystyle{IEEEtran}
\bibliography{references}

\end{document}